\documentclass{article} % For LaTeX2e
\usepackage{iclr2023_conference,times}
\usepackage{amsmath,amsfonts,bm}

\def\eqref#1{equation~\ref{#1}}
\def\1{\bm{1}}

\DeclareMathAlphabet{\mathsfit}{\encodingdefault}{\sfdefault}{m}{sl}
\SetMathAlphabet{\mathsfit}{bold}{\encodingdefault}{\sfdefault}{bx}{n}

\def\gG{{\mathcal{G}}}

\usepackage{hyperref}
\usepackage{url}

\usepackage[pdftex]{graphicx}
\usepackage{amsmath}
\usepackage{amsfonts}
\usepackage{bm}
\usepackage[linesnumbered,ruled,vlined]{algorithm2e}
\usepackage[utf8]{inputenc} % allow utf-8 input
\usepackage[T1]{fontenc}   % use 8-bit T1 fonts
\usepackage{booktabs}      % professional-quality tables
\usepackage{multirow}
\usepackage{amsfonts}      % blackboard math symbols
\usepackage{nicefrac}      % compact symbols for 1/2, etc.
\usepackage{microtype}     % microtypography
\usepackage{xcolor}
\usepackage{siunitx}
\usepackage[inline]{enumitem}   

\usepackage{wrapfig}
\usepackage{subcaption}
\usepackage{afterpage}
\usepackage{tikz}
\usepackage{pgfplots}
\pgfplotsset{compat=1.17}

\usetikzlibrary{positioning} 
\usetikzlibrary{calc}
\usetikzlibrary{pgfplots.groupplots}
\usetikzlibrary{external}
\SetKwInput{KwInput}{Input}                % Set the Input
\SetKwInput{KwOutput}{Output}              % set the Output

\usepackage[acronym,nohypertypes={acronym,notation}]{glossaries}
\newacronym{gns}{GNS}{Graph Network Simulator}
\newacronym{mgn}{MGN}{MeshGraphNet}
\newacronym{mpc}{MPC}{Model Predictive Control}
\newacronym{mbrl}{MBRL}{Model-Based Reinforcement Learning}
\newacronym{gnn}{GNN}{Graph Neural Network}
\newacronym{sofa}{SOFA}{Simulation Open Framework Architecture}
\newacronym{mpn}{MPN}{Message Passing Network}
\newacronym{mlp}{MLP}{Multilayer Perceptron}
\newacronym{cnn}{CNN}{Convoluational Neural Network}
\newacronym{mse}{MSE}{Mean Squared Error}
\newacronym{iou}{IoU}{Intersection over Union}
\newacronym{ggns}{GGNS}{Grounding Graph Network Simulator}
\newcommand{\model}{\gls{ggns} }
\title{Grounding Graph Network Simulators using Physical Sensor Observations}
\author{%
Jonas Linkerhägner$^1$\thanks{correspondence to \texttt{jonas.linkerhaegner@alumni.kit.edu}}
\And Niklas Freymuth$^1$
\And Paul Maria Scheikl$^{1,2}$
\And Franziska Mathis-Ullrich$^{1,2}$
\And Gerhard Neumann$^1$
\AND
$^1$Institute for Anthropomatics and Robotics,\\ Karlsruhe Institute of Technology, Karlsruhe, Germany
\AND
$^2$Department Artificial Intelligence in Biomedical Engineering,\\ Friedrich-Alexander-University Erlangen-Nürnberg, Erlangen, Germany
}

\iclrfinalcopy % Uncomment for camera-ready version, but NOT for submission.
\begin{document}

\maketitle

\begin{abstract}
Physical simulations that accurately model reality are crucial for many engineering disciplines such as mechanical engineering and robotic motion planning.
In recent years, learned Graph Network Simulators produced accurate mesh-based simulations while requiring only a fraction of the computational cost of traditional simulators. 
Yet, the resulting predictors are confined to learning from data generated by existing mesh-based simulators and thus cannot include real world sensory information such as point cloud data. 
As these predictors have to simulate complex physical systems from only an initial state, they exhibit a high error accumulation for long-term predictions.
In this work, we integrate sensory information to \textit{ground} Graph Network Simulators on real world observations.  
In particular, we predict the mesh state of deformable objects by utilizing point cloud data.
The resulting model allows for accurate predictions over longer time horizons, even under uncertainties in the simulation, such as unknown material properties. 
Since point clouds are usually not available for every time step, especially in online settings, we employ an imputation-based model.  
The model can make use of such additional information only when provided, and resorts to a standard Graph Network Simulator, otherwise.
We experimentally validate our approach on a suite of prediction tasks for mesh-based interactions between soft and rigid bodies. 
Our method results in utilization of additional point cloud information to accurately predict stable simulations where existing Graph Network Simulators fail. 

\end{abstract}
\section{Introduction}
Mesh-based simulation of complex physical systems lies at the heart of many fields in numerical science and engineering~\citep{liu2022eighty, reddy2019introduction, rao2017finite, sabat2021history}. 
Applications include structural mechanics~\citep{zienkiewicz2005finite, stanova2015finite}, electromagnetics~\citep{jin2015finite, xiao2022fast, coggon1971electromagnetic}, fluid dynamics~\citep{chung1978finite, zawawi2018review, long2021coupling} and biomedical engineering~\citep{van2006application, soro2018finite}, most of which traditionally depend on highly specialized task-dependent simulators.
Recent advancements in deep learning brought rise to more general learned dynamic models such as \glspl{gns}~\citep{sanchezgonzalez2018graph, sanchezgonzalez2020learning, pfaff2020learning}.
\glspl{gns} learn to predict the dynamics of a system from data by encoding the system state as a graph and then iteratively computing the dynamics for every node in the graph with a \gls{gnn}~\citep{scarselli2009the, battaglia2018relational, wu2020comprehensive}.
Recent extensions include long-term fluid flow predictions~\citep{han2022predicting} and dynamics on different scales~\citep{fortunato2022multiscale}. 
Yet, these approaches assume full knowledge of the initial system state, making them ill-suited for applications like model-predictive control~\citep{camacho2013model, schwenzer2021review} and model-based Reinforcement Learning~\citep{polydoros2017survey, moerland2020model} where accurate predictions must be made based on partial initial states and observations.

In this work, we present \glspl{ggns}, a new class of \gls{gns} that can process sensory information as input to \textit{ground} predictions in the scene observations.
More precisely, we extend the graph of the current system state with point cloud data before predicting the system dynamics from it. 
Since point clouds do not provide correspondences over time, it is difficult to learn dynamics from point clouds alone. 
Thus, we use mesh-based data to learn the general system dynamics and utilize point clouds to correct the predictions. 
As the sensory data is not always available, particularly not for future predictions, our architecture is trained with imputed point clouds, i.e., for each time step the model receives point clouds only with a certain probability. 
This training scheme allows the model to efficiently integrate the additional information whenever provided.
During inference, the model iteratively predicts the next system state, using point clouds whenever available to greatly improve the simulation quality, especially for simulations with incomplete initial state information.
Furthermore, our architecture addresses a critical research topic for \glspl{gns} by alleviating common challenges such as \textit{drift} and error accumulation during long-term predictions.

\begin{figure}
    \vspace{-0.5cm}
    \centering
	\includegraphics[width=0.90\textwidth]{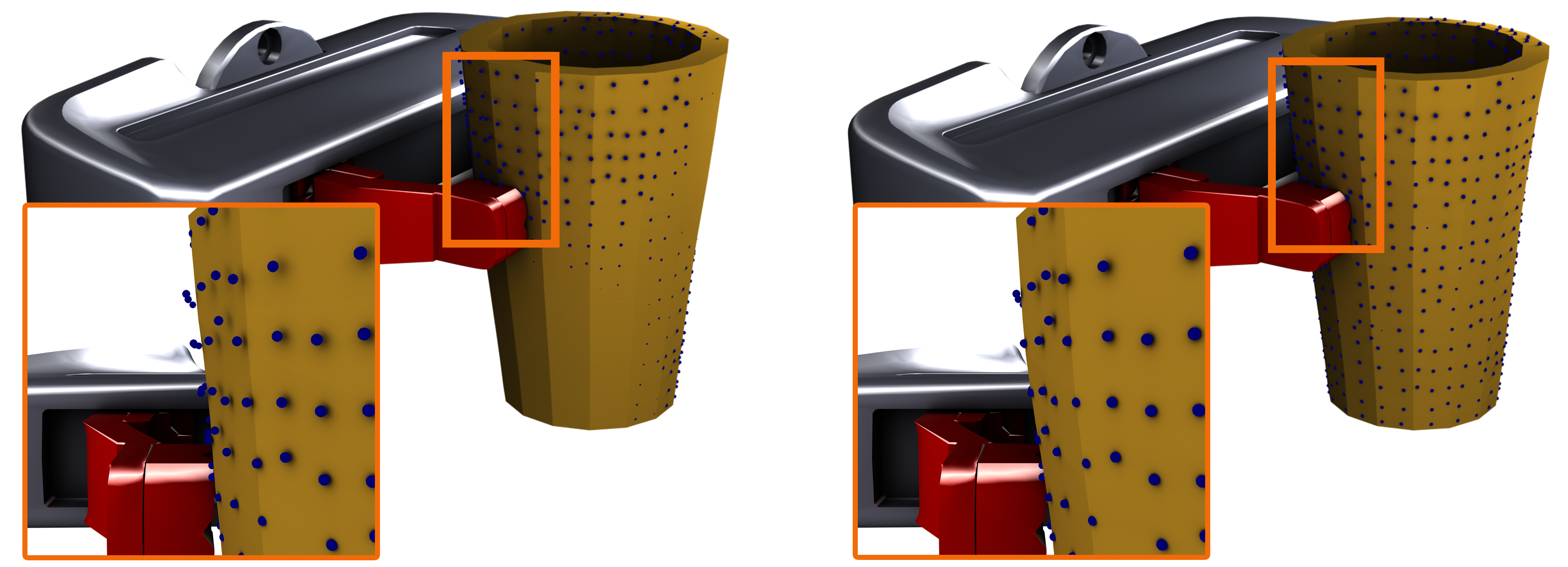}
	\caption{A robot's end-effector (grey, red) grasps a $3$-dimensional deformable cavity.
	The robot maintains an internal simulated prediction of the cavity (orange) for two consecutive simulation steps (left, right).
	This prediction can deviate from the true state of the cavity over time due to an accumulation of error. 
	However, the true cavity state can infrequently be observed from point cloud data (blue), which the model can use to correct its prediction.  
	% be integrated into the simulator to correct the prediction. 
	Here, the point cloud is used to contract the simulated cavity at the bottom and extend it at the top, causing the points to better align with the mesh surface. 
	We repeat the point cloud from the earlier simulation step in both images for clarity.
	}
    \label{fig:figure_one}
    \vspace{-0.25cm}
\end{figure}

As a practical example, consider a robot grasping a deformable object.
For optimal planning of the grasp, the robot needs to model the state of the deformable object over time and predict the influence of interactions between object and gripper. 
This prediction not only depends on the initial shape of the object, but also on the forces the robot applies, the kind of material to grasp and external factors such as the temperature, making it difficult to accurately predict how the material will deform over time. 
However, once the robot starts deforming the object, it may easily observe the deformations in the form of e.g., point clouds. 
These observations can then be integrated into the state prediction, i.e., they can \textit{ground} the simulation whenever new information becomes available.
An example is given in Figure~\ref{fig:figure_one}.
Such observation-aided prediction is similar in nature to e.g., Kalman Filters~\citep{kalman1960filter, jazwinski1970stochastic, becker2019recurrent} as the belief of the system state is updated based on partial observations about the system. 
However, while Kalman Filters explicitly integrate novel information into the belief in a mathematical fashion, we instead simply provide this information to a learned model as additional unstructured sensor input.
% 
% As a side note, mpc can usually try different actions from the current state. Can we also easily do that by e.g., modifying the velocity features of the actuator in the deforming plate task \textit{during runtime}? If not, that may be another experiment to consider.

We evaluate \model on a suite of $2$d and $3$d deformation prediction tasks created in the \gls{sofa}~\citep{faure2012sofa}. 
% These tasks include settings with both fully and partially observable initial system states.
Comparing our approach to an existing \gls{gns}~\citep{pfaff2020learning}, we find that adding sensory information in the form of point clouds to our model improves the simulation quality for all tasks. 
% This benefit is most apparent when the initial system state is not fully known, e.g., when the exact physical properties of the material are unknown. 
We investigate this behavior through extensive ablation studies, showing the importance of different parameter choices and design decisions.
%To facilitate research in this area, 
Code and data can be found under \url{https://github.com/jlinki/GGNS}.

% To summarize, we (i) extend the \gls{gns} framework to include sensory information to \textit{ground} predicted simulations in observations of the system state, (ii) propose a simple but effective imputation training scheme that allows our model to utilize sensory information whenever available and to resort to a regular \gls{gns} otherwise, (iii) construct and experiment on different deformation prediction tasks and find that the inclusion of sensory information improves performance in all settings, and that it is particularly crucial when the initial system state is not fully known.

Our list of contributions is as follows:
% should be 3 distinct things, 1 of which is the experiments
\begin{enumerate*}[label=(\Roman*),itemjoin={{ }}]
    \item We extend the \gls{gns} framework to include sensory information to \textit{ground} predicted simulations in observations of the system state, allowing for accurate predictions of the full simulation.
    \item We propose a simple but effective imputation training scheme that naturally integrates sensory information to \glspl{gns} whenever available without substantially increasing training cost or model complexity.
    \item We construct and experiment on different deformation prediction tasks and find that the inclusion of sensory information improves performance in all settings, and that it is particularly crucial when the initial system state is not fully known.
\end{enumerate*}

\section{Related Work}
\label{sec:related_work}

\textbf{\emph{Learned Physics Simulation.}}
In recent years there has been a steady increase in research concerning deep learning for physical simulations.
Early work in physical reasoning aims at teaching systems to understand physical relations on N-body systems~\citep{battaglia2016interaction} and deformable objects~\citep{mrowca2018flexible}.
A more direct approach is to instead train a learnable simulator from data provided by some existing ground truth simulator. Here,
\glspl{cnn} have been extensively studied for fluid flow simulation~\citep{tompson2016accelerating, chu2017data-driven, ummenhofer2020lagrangian, kim19deep, xie2018tempoGAN}
% added last 2 citations from adversarial sentence below
and aerodynamic flow fields~\citep{guo2016convolutional, zhang2018application, bhatnagar2019prediction}. 
% These methods have also been combined with Generative Adversarial Networks~\citep{goodfellow2014generative} to generate fluid flow simulations in an adversarial fashion~\citep{kim19deep, xie2018tempoGAN}.
% cut this to save some space
Further approaches use standard neural networks for liquid splash simulations~\citep{um2018liquid} and latent space physics simulation~\citep{wiewel2019latent}.
%Advantages of such learned learned physics simulators include that they 
Such learned physics simulators are considerably faster than their ground-truth counterparts, and that they are usually fully differentiable.
Thus, they have been applied to model-based Reinforcement Learning~\citep{mora2021pods} and for Inverse Design problems~\citep{baque2018geodesic, durasov2021debosh, allen2022physical}
%, i.e., to design objects to exhibit desired properties based on simulation
.

\textbf{\emph{Graph Network Simulators.}}
Graph Network Simulators (GNS)~\citep{sanchezgonzalez2020learning} are a special case of learned physics simulators that utilize \glspl{gnn}~\citep{scarselli2009the} to efficiently encode the graph-like structure of many physical problems.
They have found wide-spread application in calculating atomic forces~\citep{hu2021forcenet}, particle-based simulations~\citep{li2019learning, sanchezgonzalez2020learning} and mesh-based simulations~\citep{pfaff2020learning, weng2021graph-based, han2022predicting, fortunato2022multiscale, anonymous2022graph}.
Other works in this field directly solve partial differential equations~\citep{alet2019graph}, and integrates explicit domain knowledge into the learned simulator to improve the predictions~\citep{belbute-peres2020combining, li2021accelerating, li2022graph}.
Similarly, \glspl{cnn} have been used to predict particle masses from images to subsequently simulate physical systems with a \gls{gnn} ~\citep{li2020visual} via visual grounding.
This approach assumes access to a series of images to predict particles and their behavior, whereas \model integrates sensor observations into an existing mesh-based simulation.
The work most closely related to our research is \gls{mgn}~\citep{pfaff2020learning}, which combines a graph-based encoding of the system state with the next-step prediction of dynamic quantities to produce realistic predictions of mesh-based simulations. 

\textbf{\emph{Simulation from Observation.}}
Another variant of learned physics simulation is simulation from observation.
Learning directly from observations instead of a ground truth simulator requires less expert knowledge for the design of the simulator and is more applicable to real-world scenarios.
Different approaches exist for this type of simulation, including Physical reasoning~\citep{li2020visual} and particle-based simulation~\citep{martinkus2020scalable}.
Point clouds have been used in \gls{cnn}-based simulation~\citep{watters2017visual, wand2018dynamic}, and combined with PointNet~\citep{qi2016pointnet, ruizhongtai2017pointnet} to predict object deformations purely from observational data~\citep{park2021flexible}.
Further approaches make use of \glspl{gnn} to predict object relations \citep{fetaya2018neural} and future frames in a point cloud sequence \citep{gomes2021spatio-temporal}.

\textbf{\emph{Simulation of Deformable Objects.}}
Simulating deformable objects is crucial for many applications such as robotic manipulation tasks~\citep{sanchez2018robotic}. 
Yet, recent approaches do not take explicit deformation into account~\citep{jan2018sim-to-real}, or only consider highly simplified geometries such as ropes~\citep{sundaresan2020learning} or a square piece of cloth~\citep{yilin2020learning, lin2020softgym, lin2022learning}. 
One reason for this is the high computational cost of existing simulators, which may be alleviated by fast and accurate learned simulators~\citep{pfaff2020learning, weng2021graph-based}.
Another recent work trains the parameters of a differentiable simulator to align its simulations with real-world observations of deformable objects based on point cloud information ~\citep{sundaresan2022diffcloud}. 
In this work, we instead utilize point cloud information to improve upon existing mesh-based \glspl{gns} in settings where additional point cloud data is available.
% produce realistic data by comparing its simulations with real-world data using point cloud information~\citep{sundaresan2022diffcloud}. 

% \todo{
% Geri:
% I would add the latest work from Jeannette Bohg, for example her IROS paper. They use differentiable simulators there. Pro: very efficient. Cons: Limited as it can only model effects that are modelled by the simulator. \cite{sundaresan2022diffcloud}
% }

\section{Foundations}

\subsection{Message Passing Network}
Let $\displaystyle \gG = (\mathbf{V}, \mathbf{E}, \mathbf{X}_\mathbf{V}, \mathbf{X}_\mathbf{E})$ be a directed graph with nodes $\mathbf{V}$, edges $\mathbf{E} \subseteq \mathbf{V}\times \mathbf{V}$, node features $\mathbf{X}_\mathbf{V}: \mathbf{V}\rightarrow \mathbb{R}^{d_\mathbf{V}}$ of dimension $d_\mathbf{V}$ and edge features $ \mathbf{X}_\mathbf{E}: \mathbf{E}\rightarrow \mathbb{R}^{d_\mathbf{E}}$ of dimension $d_\mathbf{E}$. 
A \gls{mpn} \citep{sanchezgonzalez2020learning, pfaff2020learning} is a \gls{gnn} consisting of $L$ \textit{Message Passing Blocks} that receives the graph $\gG$ as input and outputs a learned representation for each node $\mathbf{V}$ and edge $\mathbf{E}$. 
Each block $l$ computes updated features for all nodes $v\in \mathbf{V}$ and edges $e\in\mathbf{E}$ as
\begin{align*}
\textbf{x}^{l+1}_{e} &= f^{l}_{\mathbf{E}}(\textbf{x}^{l}_v, \textbf{x}^{l}_u, \textbf{x}^{l}_{e}), \textrm{ with } e = (u, v)  \textrm{ and }
\textbf{x}^{l+1}_{v} = f^{l}_{\mathbf{V}}(\textbf{x}^{l}_{v}, \bigoplus_{\{e=(v,u)\in \mathbf{E}\}} \textbf{x}^{l+1}_{e})\text{,}
\end{align*}
where $\textbf{x}_v^0$ and $\textbf{x}_e^0$ are embeddings of the initial node and edge features of $\gG$ and $\oplus$ is a permutation-invariant aggregation such as a sum, max, or mean operator. 
Furthermore, each $f^l_\cdot$ is a learned function that is generally parameterized as a simple \gls{mlp}.
% Previous work \citep{pfaff2020learning} decomposes these learned functions to explicitly represent two different kinds of edges, in their case mesh and world edges.
% More concretely, they assume an edge partition $\mathbf{E} = \mathbf{E}_1 \dot{\cup} \mathbf{E}_2$ and separate edge update functions $f^l_{\mathbf{E}_1}$ and $f^l_{\mathbf{E}_2}$. 
% The edge-aggregation for the node update is then computed by aggregating the latent features of both types of edges separately and concatenating the result. 
% Our method, however, omits this explicit representation in favor of a simple one-hot encoding of the type of input edge, because we did not find any significant advantages of explicit partitioning of edge types over one-hot encodings for our tasks. For more details, see Appendix~\ref{app_sec:ablations}.
% A comparison to the explicit representation of different types of edges is conducted in our experiments.

\subsection{Graph Network Simulator}
\glspl{gns} simulate a system's dynamics by repeatedly applying the following three steps.
First, they encode the system state $\mathcal{S}$ in a graph $\gG$.
% using some encoding $\text{enc}(\mathcal{S})=\gG$
If the system state is given as e.g., a triangular or tetrahedral mesh $\mathcal{M}$ of the underlying entities, this graph is naturally constructed by using the nodes of $\mathcal{M}$ as nodes of the graph, and the connection between these nodes as edges.
The node and edge features $\mathbf{X}_\mathbf{V}, \mathbf{X}_\mathbf{E}$ can be constructed based on the concrete simulation. 
In general, encoding purely \textit{relative} properties such as relative distances and velocities per edge rather than absolute positions per node have been shown to greatly improve training speed and generalization \citep{sanchezgonzalez2020learning}. 
Next, the encoded graph $\gG$ is used as input for a learned \gls{mpn},
% $\text{mpn}(\gG)=\{x_v^L\}$
which computes final latent representations $x_v^L$ for each node $v\in\mathbf{V}$. 
These latent representations are interpreted as (potentially higher-order) derivatives of dynamic quantities, which are used by a simple forward-Euler integrator to derive an updated system state $\mathcal{S}'$.
%via some decoder $\mathcal{S}' = \text{dec}(S, \{x_v^L\})$
% This is usually done by interpreting the final latent representation of each predicted node as a (potentially higher-order) derivative of its dynamic properties and using a simple forward-Euler integrator to update said properties to derive the updated system state $\mathcal{S}'$.
Note that for some tasks, only a fraction of mesh nodes need to be predicted, as the others are either fixed or belong to a known entity such as a gripper or collider. In this case, only the latent representations of the nodes with otherwise unknown dynamics are used.

\glspl{gns} are trained on a node-wise next-step \gls{mse} objective, i.e., they minimize the $1$-step prediction error of the next system state to that of a given ground truth trajectory.
During inference, simulations over potentially hundreds of steps can be generated by iteratively repeating the above-mentioned steps, using the updated dynamics of one step as the input for the next.
We note that the model does not predict the movement of fixed entities such as e.g., a collider, which is instead assumed to be known and combined with the model's prediction about the unknown parts of the system.
Due to this iterative dependence on previous outputs, the model is prone to error accumulation. A common strategy to tackle this limitation is to apply additional noise to the dynamic variables of the system for each training step \citep{sanchezgonzalez2020learning, pfaff2020learning}.
Intuitively, adding training noise acts as a form of data augmentation that allows the learned model to compensate for small prediction errors over time. This kind of error-compensating next-step prediction leads to plausible and visually realistic predictions. However, the resulting predictions can be arbitrarily inaccurate with respect to the true dynamics of the system, since the model has no reference for its simulation other than some potentially incomplete initial state $\mathcal{S}_0$.

% During inference, the resulting models can be used to generate stable simulations for hundreds of steps by iteratively repeating the above-mentioned steps.
% That is, given a dataset consisting of series of system states $\mathcal{S}_0, \dots, \mathcal{S}_T$, they are tasked to predict updates that minimize the error to the actual next system state, i.e., to minimize
% $$
% \mathcal{L} = \mathbb{E}_{\mathcal{S}_t, \mathcal{S}_{t+1}}\left[||\mathcal{S}_{t+1}-\text{dec}(S, \text{mpn}(\text{enc}(\mathcal{S})))||^2 \right]
% $$
% probably no artificial mathiness needed here...

% \begin{enumerate}
%     \item Briefly define a (heterogeneous) graph (with features on nodes and edges)
%     \item Define and explain GNN blueprint, message passing network. 
%     \item Explain mgn. Briefly go into relative encoding, training noise and next-step objective
%     \item Maybe something related to point cloud observations
% \end{enumerate}
\section{Grounding Graph Network Simulator}
\begin{figure}
    \vspace{-0.5cm}
    \centering
	\includegraphics[width=0.9\textwidth]{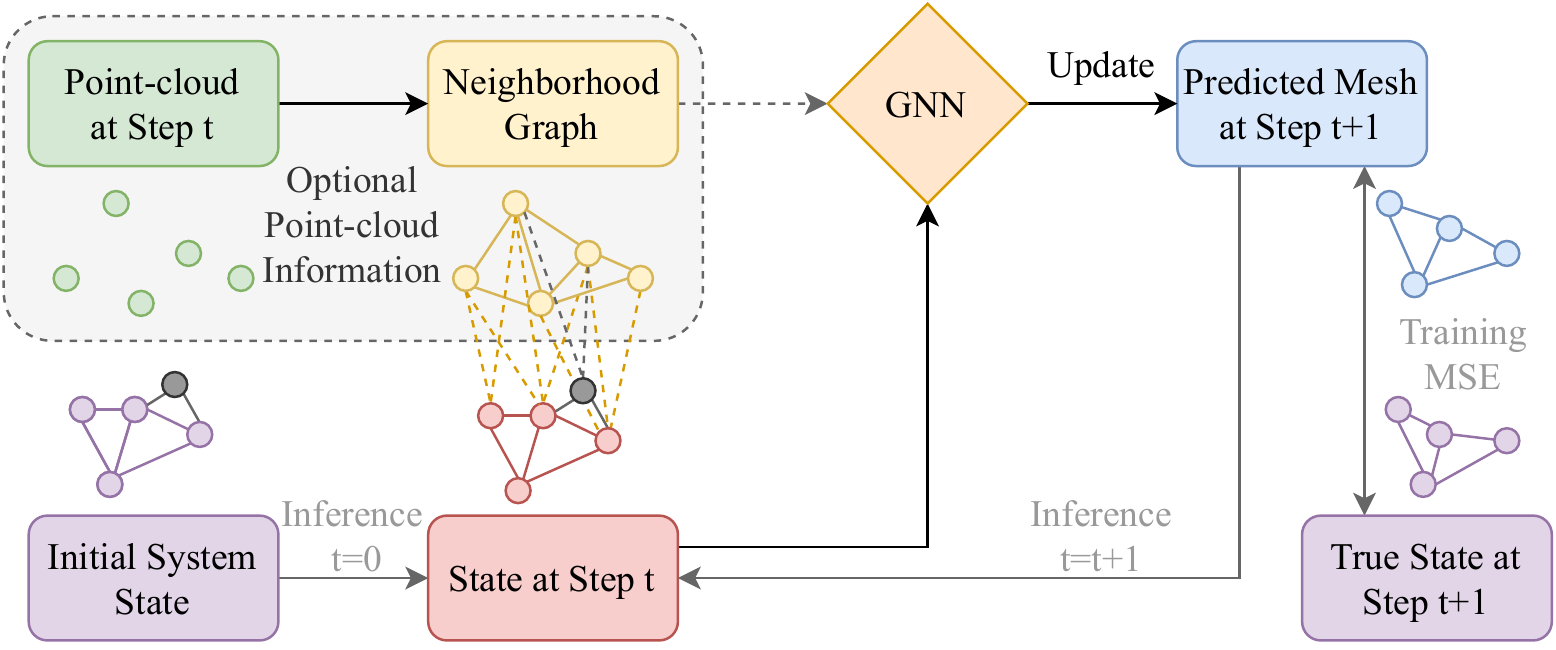}
	\caption{Schematic of \model.
	Given a system state $\mathcal{S}_t$ (red) and optional point cloud observations (dashed box), a \gls{gnn} (orange) predicts how the system $\mathcal{S}_{t+1}$ will look like at the next step (blue). 
	For object deformation tasks, the state can include boundary conditions (gray) such as colliders or walls.
	When provided, the point cloud (green) is transformed into a neighborhood graph (yellow) in the same coordinate system as the mesh, connecting each point in the point cloud to the nearest mesh nodes. 
	The model is trained to predict the next system state based on a true state (purple) provided by a ground truth simulator.
	During inference, the model iteratively predicts updates from a potentially incomplete initial system state (purple), using additional point cloud observations when available.}
    \label{fig:overall_architecture}
    \vspace{-0.25cm}
\end{figure}

Our approach combines recent advances in graph-based learned physics simulation with additional partial observations of the system state to generate highly accurate simulations from incomplete initial states.
To this end, we extend the existing \gls{gns} framework to naturally and efficiently integrate auxiliary point cloud data whenever available. This auxiliary information \textit{grounds} the predictions of the model in an observation of the true system state, guiding it towards predictions that not only look realistic but also closely match the actual dynamics of the system.
Figure~\ref{fig:overall_architecture} illustrates an overview of our approach. A more detailed description of the GNN-part of the method is found in Appendix~\ref{app_sec:detailed_model}.

\subsection{Point Clouds and Neighborhood Graphs}
\label{ssec:method:point_clouds}
% {The main challenge when using point clouds is that there are no point correspondences over time. 
% This means the model needs to figure out which point of the point cloud belongs to which vertex in the mesh to do the correction of the mesh nodes in order to ground the simulation.
In order to utilize point-based data in addition to meshes we first have to transfer both into a common graph. Following previous work \citep{sanchezgonzalez2018graph}, we do this by creating a neighborhood graph based on spatial proximity. Given a graph $\gG = (\mathbf{V}, \mathbf{E}, \mathbf{X}_\mathbf{V}, \mathbf{X}_\mathbf{E})$ that encodes a predicted system state and a point cloud observation $\mathcal{P}=\{p_1, \dots, p_N\}$, $p_j\in\mathbb{R}^d$ of the true system state, we set $\mathbf{V}'=\mathbf{V}\cup\mathcal{P}$ and
$$
\mathbf{E}' = \mathbf{E} 
\cup \{(p_i, p_j)\in \mathcal{P}^2 | \text{d}(p_i, p_j)\leq r_\mathcal{P}\} \cup \{(v, p), (p, v) | v\in \mathbf{V}, p\in\mathcal{P}, d(p, v)\leq r_\mathcal{S}\}\text{.}
$$
Here, $d$ is some distance measure, usually the euclidean distance, and $r_\mathcal{P}$ and $r_\mathcal{S}$ are task-specific neighborhood radii.
The corresponding features $\mathbf{X}_{\mathbf{V}'}$, $\mathbf{X}_{\mathbf{E}'}$ of the added nodes and edges in $\mathbf{V}'$ and $\mathbf{E}'$ depend on the concrete task. 
The different node and edge types are one-hot encoded into their respective features to allow the model to differentiate between them. 
Similar to the original features, information can be encoded in a relative fashion in the form of edge features to aid generalization. More concretely, we encode relative distances in world space along all edges, additionally adding mesh-space distances for edges between two mesh nodes.
This connectivity is slightly different from \gls{mgn} \citep{pfaff2020learning}, which make use of additional \textit{world edges} between mesh-nodes by creating a similar radius-based neighborhood graph for the mesh nodes in world space. 
% We find that adding such world edges does not improve our method, instead we encode mesh distances in world space into the existing mesh edges \todo{rephrase to not mention results too early}. For completeness, we compare to \gls{mgn} both with and without world edges in our experiments.

% Intuitively, each predicted system node $v\in\mathbf{V}$ will be connected to close-by points of a point cloud which it can use to reason about the true system state, potentially correcting its own wrong behavior. 
% An example of is given in Figure \ref{fig:figure_one} \todo{refer back to figure one here, assuming that it actually is a good example.}. Here, the system state consists of a mesh and a collider, and the point cloud consists of points sampled from the meshed object. \todo{maybe another sentence for intuition.}

\subsection{Imputation-based training and inference}
For most realistic applications, point clouds are typically not available at each time step during inference. For example, we may have access to observed point clouds from the previous $k$ time steps and want to use them to infer the state of the system in the future.
We adapt our model to this constraint by employing an imputation-based training scheme.
Our model still uses a single \gls{gnn}, but we now randomly replace the graph $\gG$ of $\mathcal{S}$ with the corresponding extended graph $\gG' = (\mathbf{V}', \mathbf{E}', \mathbf{X}_{\mathbf{V}'}, \mathbf{X}_{\mathbf{E}'})$ with equal probability during training.
% a probability of $k=0.5$
In both cases, the model is only trained to predict the system dynamics for the original nodes $\mathbf{V}$. 
Intuitively, this allows each system node $v\in\mathbf{V}$ to utilize the additional information of close-by points of a point cloud when available, while at the same time forcing it to also make sensible predictions when there is no additional information. 
During inference, we construct $\gG'$ from the (predicted) system state $\mathcal{S}$ and a corresponding observed point cloud $\mathcal{P}$  of the true object whenever available and use $\gG$ otherwise. 
This enables the model to reason about the true system state that is observed via $\mathcal{P}$, adapting its prediction to the otherwise unknown behavior of the system. This \textit{grounding} of the prediction also alleviates common errors of \gls{gns} such as drift and more generally error accumulation.
An example can be seen in Figure~\ref{fig:figure_one}.
Here, the system state consists of a predicted mesh and a gripper, and the point cloud consists of points sampled from the true object. The mismatch between point cloud and predicted mesh indicates the prediction error, and the model uses this additional information to correct the current state estimate.
Similar figures for the other two tasks can be found in Appendix~\ref{app_sec:qualitative_results}.
% Intuitively, each predicted system node $v\in\mathbf{V}$ will be connected to close-by points of a point cloud in the extended graph, which it can use to reason about the true system state to potentially correct its own wrong behavior. An example of is given in Figure \ref{fig:figure_one} \todo{refer back to figure one here, assuming that it actually is a good example.}. 

We compare this simple imputation-based method to another training scheme in our experiments, which we call \gls{ggns}+LSTM. Here we use an LSTM~\citep{hochreiter1997long} layer on the node output features of the \gls{gnn} to explicitly include recurrency into the model. 
This modification allows information such as the material properties to be inferred and propagated over time, which can be utilized to improve the predictions in time steps without point clouds. 
The resulting model is trained on the same 1-step prediction loss and also uses training noise to generate stable rollouts during inference.
However, it is significantly more costly to train, as it makes use of backpropagation in time to compute the gradients for the recurrency.
We find experimentally that this recurrent model performs worse than the imputation-based method.
An explanation for this is that the potential benefit of propagating information over time is offset by the additional training and model complexity, especially with respect to the next-step prediction objective.
For this reason, \model relies on this simple but effective imputation-based approach.
% \todo{Geri: Yes, already say something about the results. Does it work better or worse?} 

% \begin{enumerate}
%     \item Rename section to the name of our approach.    
%     \item Use a nice Figure 2 that shows a schematic of the approach
%     \item Describe in detail what we are doing with our approach, including the input construction and encoding. Highlight novelty and use cases (again). Could consider adding a plot of e.g., the neighborhood graph here for the deforming plate task here
%     \item Motivate (again) why we do this, i.e., how it helps with shift and accumulating errors by "grounding" the model geometry in something
%     \item Explain imputation approach and how the network is trained to cope with this. Mention (again) that we have experiments and ablations supporting our claims
% \end{enumerate}

\section{Experiments}
\label{sec:experiments}

\begin{figure}
    \vspace{-0.5cm}
    \centering
	\includegraphics[width=0.72\textwidth]{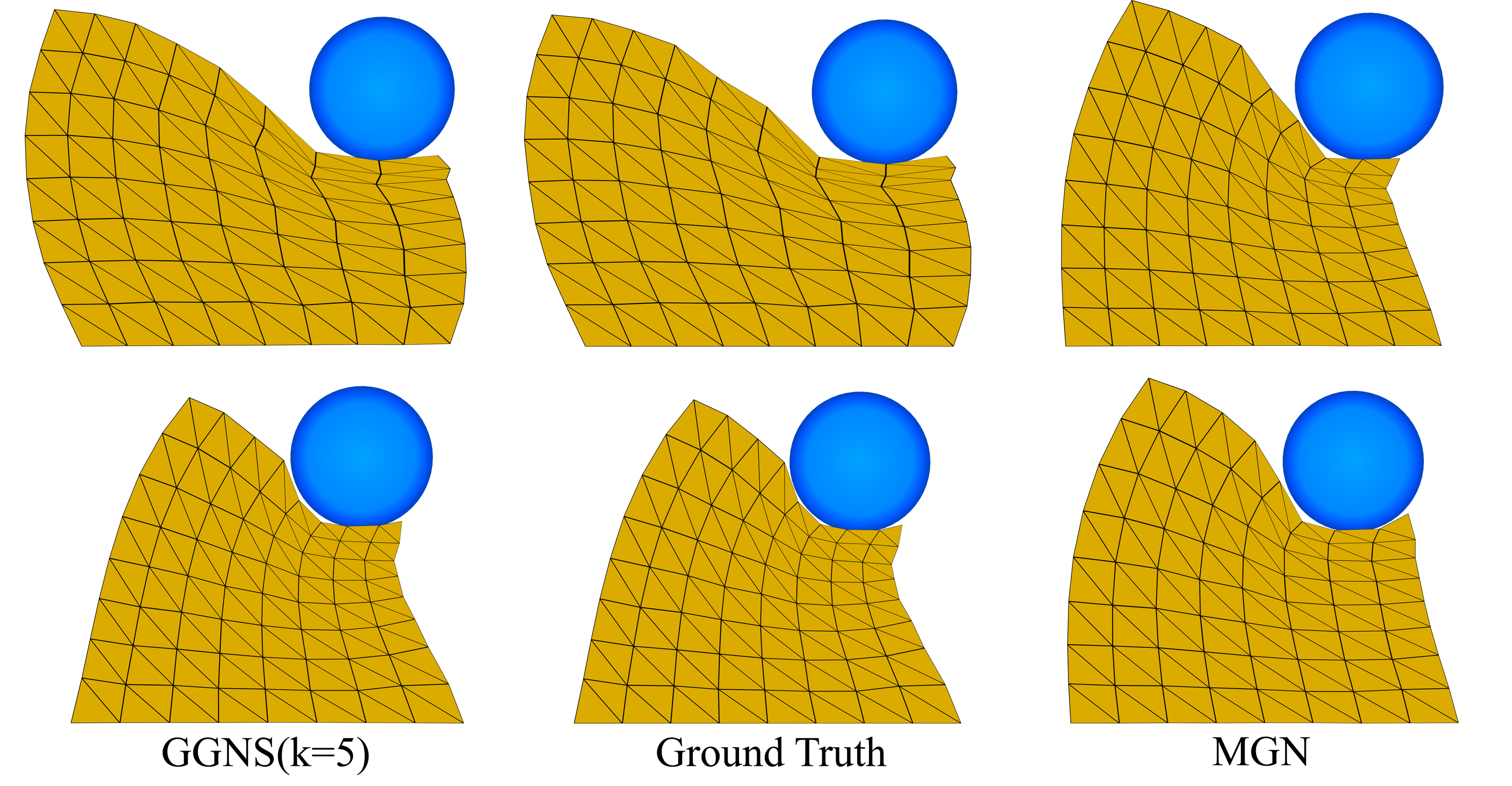}
	\caption{Final simulated meshes ($t=50$) for \model ($k=5$) (left), the ground truth simulation (middle) and \gls{mgn} (right) for $2$ test rollouts with different material properties for the Deformable Plate task. Our model closely matches the ground truth simulations for both materials, while \gls{mgn} predicts the same material every time.}
    \label{fig:trapez_qualitative}
    \vspace{-0.25cm}
\end{figure}

We evaluate \model on complex $2$d and $3$d mesh-based object deformation prediction tasks modelled in the \glsfirst{sofa}~\citep{faure2012sofa}. 
For each task, the true system state is given by a tetrahedral FEM mesh of a deformable object with rigid boundary conditions combined with a triangular surface mesh of a rigid collider.
The point clouds are generated by raycasting using one virtual camera for $2d$ and up to five cameras for $3d$ tasks arranged around the scene. 
More details on the generation of the point clouds are presented in Appendix~\ref{app_sec:environments}.
% and ablations on the number of cameras can be found in Appendix~\ref{app_sec:ablations}.
Additional environment-specific details, including node and edge features and dataset properties can also be found in Appendix~\ref{app_sec:environments}.
We assume that, while the initial mesh of the object is known, its material properties are not. 
We model these unknown properties via the Poisson's ratio~\citep{lim2015auxetic} $-1<\nu<0.5$, which is a scalar value describing the ratio of contraction ($\nu<0$) or expansion ($\nu >0$) under compression~\citep{mazaev2020auxetics}. 
For all datasets, we randomly assign Poisson's ratios from $\nu \in \{-0.9, 0.0, 0.49\}$ equally to all rollouts. 
%A larger ratio of $v\approx 0.5$ implies a constant object volume under deformation, as with e.g., rubber, whereas smaller ratios $v<0$ imply an \textit{auxetic} material that also exhibits lateral compression under deformation, which appears for e.g., certain mineral structures~\citep{mazaev2020auxetics}.
% A large ratio of $v\approx 0.5$ means that the volume of the object stays constant under deformation by expanding perpendicular to the applied force, as with e.g., rubber. 
% A ratio $v=0$ implies that the object is compressed when deformed by a force, such as for cork. 
% Smaller ratios $v<0$ imply an \textit{auxetic} material that also exhibits lateral compression under deformation, which appears for e.g., certain mineral structures~\citep{mazaev2020auxetics}.

We train all models on all tasks using the Adam optimizer \citep{kingma2014adam} with a learning rate of $5\times 10 ^{-4}$ and a batch size of $32$, using early stopping on a held-out validation set to save the best model iteration for each setting.
The models use a LeakyReLU activation function, five message passing blocks with $1$-layer MLPs and a latent dimension of $128$ for node and edge updates. We use a mean aggregation for the edge features and a training noise of $0.01$.
All tasks use a normalized task space of $[-1, 1]^d$.
%For all models and tasks, we normalize the task space to $[-1, 1]^d$ and add a training noise of $0.01$.
%All experiments are repeated for $10$ random seeds unless otherwise noted, and we report the mean and standard deviation of the results.
An overview of the network hyperparameters can be found in Appendix~\ref{app_sec:hyperparameters}.

\textbf{Evaluation Metrics.} 
We evaluate the performance of all trained models on $10$ different seeds per experiment.
We report the means and standard deviations of the different runs, where, for each run, we average the results over all available steps of a trajectory and over all trajectories in the test set of the respective data set.
%We report means and standard deviations of the average results over all applicable steps on held-out evaluation simulations.
%drawn from the same distribution as the training data. 
% For all experiments, we report the \textit{$1$-step error} over the predicted object mesh, i.e., we provide the model with a true system state $\mathcal{S}_t$ and compare its prediction to the true state $\mathcal{S}_{t+1}$ for each predicted mesh node. 
For all experiments, we report the \textit{full rollout loss}, where the model starts with the initial state $\mathcal{S}_0$ and predicts the states up to a final state $\mathcal{S}_T$. 
Here, we provide a point cloud to the model every $k\geq 1$ steps and resort to mesh-only prediction otherwise. 
This corresponds to a setting in which the deformation of an object is tracked with both high-frequency sensors and low-frequency cameras which provide the position of the rigid collider and point-cloud information respectively.

We also consider an application where a robot observes an object's deformation up to some point in time and then reasons about future deformations without additional point-cloud information.
For this setting, the initial system state $\mathcal{S}_0$ is provided to the model, followed by $m$ point clouds for its next $m$ predictions. 
Then, $10$ more steps are predicted without point clouds to predict a state $\mathcal{S}_{m+10}$ and and compute the corresponding \textit{$10$-step prediction loss}.
The reported losses are the average \gls{mse} over every step along the trajectory averaged over all possible rollouts.
This metric reduces to the average loss for a $m+10$-step prediction for methods that cannot make use of point cloud data, as the state $\mathcal{S}_{m+10}$ needs to be predicted from the initial $\mathcal{S}_0$.

% \begin{enumerate}
%     \item All ground truth simulations come from SOFA on custom environments.
%     \item For each environment, briefly describe \textit{why} we choose it and what makes it interesting.
%     \item Go into more detail for each environment and its intricacies (which parameters, how many points of data, ...), although this is very likely to go into the appendix later on.
% \end{enumerate}

\textbf{Baselines.} We compare to \gls{mgn}, a state-of-the-art \gls{gns}, which utilizes additional \textit{world edges} between close-by mesh nodes, but does not incorporate point cloud observations.
Comparing these world edges to  Section~\ref{ssec:method:point_clouds}, \gls{mgn} assumes an edge partition $\mathbf{E} = \mathbf{E}_1 \dot{\cup} \mathbf{E}_2$ and separate edge update functions $f^l_{\mathbf{E}_1}$ and $f^l_{\mathbf{E}_2}$. 
The edge-aggregation for the node update is then computed by aggregating the latent features of both types of edges separately and concatenating the result. 
We adopt this explicit representation of edge types for the \gls{mgn} baseline and experiment with it for \gls{ggns} in Appendix~\ref{app_sec:ablations}.
As it does not provide any significant advantages for our model, \gls{ggns} instead resorts to a simple one-hot encoding of the type of input edge for the remaining experiments.

% We consider different variants of \gls{mgn} to compare the performance of our method to that of a state-of-the-art \gls{gns}. 
% These baselines are chosen to contrast the performance of our method with that of a state-of-the-art \gls{gns}, and to explain how some of our design decisions affect results. 
 % We find that adding world edges does not improve our method, and instead encode world space in the existing mesh edges.
Additionally, we evaluate a variant of \gls{mgn} that has additional access to the underlying Poisson's ratio $v$ as a node feature, called \textit{\gls{mgn} (M)}. 
This additional information leads to a deterministic ground truth simulation w.r.t. the initial system state, and upper bounds the performance of \gls{mgn}.
We also compare to \gls{ggns}+LSTM, which integrates recurrency into our imputation technique.
Here, we investigate whether this recurrency helps the model predicting e.g., material properties over time. 
%which integrates recurrency into our imputation technique to allow the model to infer and propagate, e.g., material properties over time.
%an imputation technique that uses, namely . The explicit inclusion of recurrence into the model may help to pass information over time, especially during time steps without point cloud information.

\begin{figure}
    \centering
    \vspace{-0.5cm}
	\includegraphics[width=0.82\textwidth]{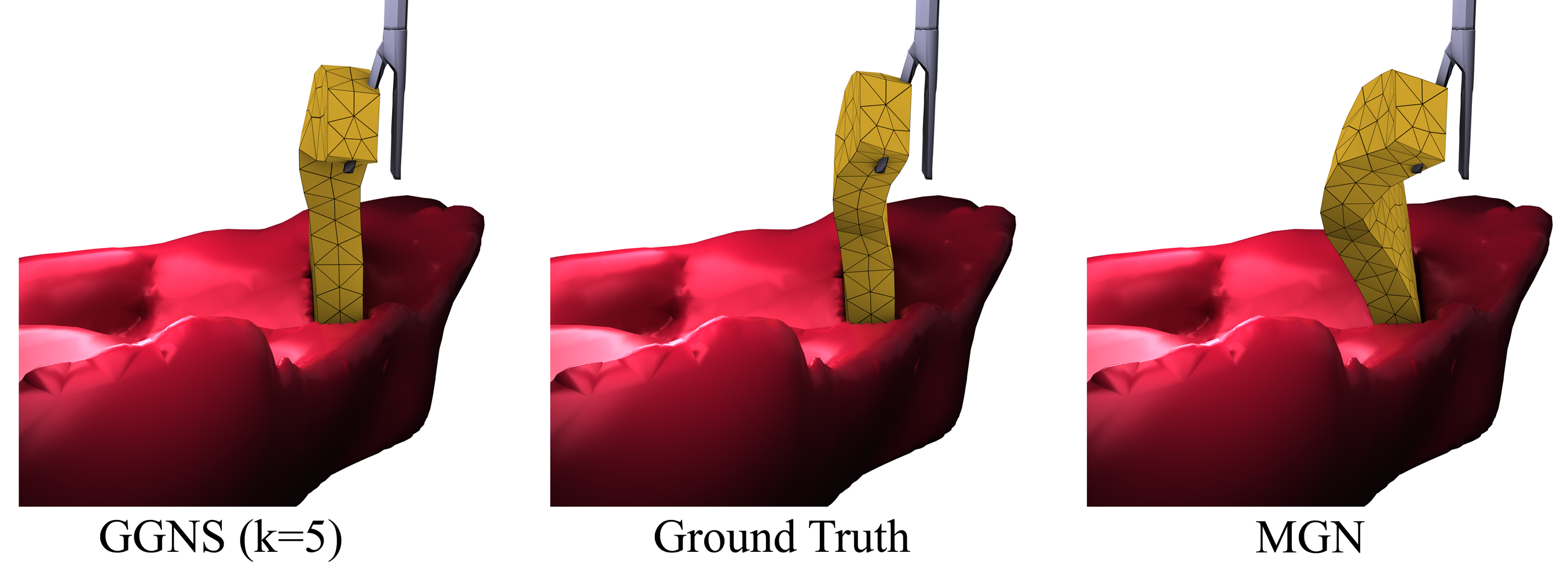}
	\caption{Visualization for a test trajectory at time step $t=70$ for \model (left), the ground truth simulation (middle) and \gls{mgn} (right). While \gls{mgn} accumulates a large prediction error over time, \model is able to utilize the additional point cloud information to stay close to the ground truth for the full length of the simulation. Different angles of this visualization can be found in Appendix~\ref{app_sec:qualitative_results}.}
    \label{fig:tissue_qualitative}
    \vspace{-0.25cm}
\end{figure}

As a point cloud based baseline, we use a non-learned method to 
%dditionally, we consider a non-learned method, which 
directly generate a mesh from the point cloud of each time step. 
We voxel-subsample the point cloud so that we observe approximately the same number of points as nodes in the ground truth mesh and then use \textit{Alpha Shapes} \citep{akkiraju1995alpha} to create a (potentially non-convex) mesh for this time step.
This baseline shows how much information can be directly inferred from just the point cloud information.
%we use the point cloud to calculate the convex hull, which is then transformed to a mesh via Delaunay triangulation. 
%To account for non-convex shapes,
%To showcase how much information can be directly inferred from just the point cloud observation, we create a mesh from the point of every time step. 
%To account for the deformation, we need a non-convex mesh generation method. 
%To this end, we voxel-subsample the point cloud so that we get approximately the same number of points as nodes in the ground truth mesh and then use so-called \textit{Alpha Shapes}  for the mesh generation. 
% We could also consider an additional CNN-based baseline, though we should probably wait until the rebuttal if one of the reviewers complains about that.

\textbf{Deformable Plate.}
%As the first task, w
We consider a family of $2$-dimensional trapezoids that are deformed by a circular collider with constant velocity. 
Besides the trapezoidal shapes, diversity in the dataset is introduced by varying the size and starting positions of the collider. 
For this task, we additionally consider the \gls{iou} between the predicted and the ground truth mesh as an evaluation metric. 
We find that this metric is less sensitive to individual mesh nodes and that it instead measures how well the predicted object shape matches that of the real system state.
We use a total of $675/135/135$ trajectories for our training, validation and test sets. Each trajectory consist of $T=50$ timesteps.

\textbf{Tissue Manipulation.}
An important application for the prediction of deformable objects is medical robotics. 
%For this reason, the use of our method can be useful in this context.  
We simulate a robot-assisted surgery scenario where a piece of tissue is deformed by a solid gripper.
Varying the direction of the gripper's motion and its gripping position on the tissue results in additional diversity.
Here, $600/120/120$ trajectories are used, each of which is rolled out for $T=100$ timesteps.
This task is visualized in Figure \ref{fig:tissue_qualitative}.

\textbf{Cavity Grasping.}
Robotic manipulation of deformable objects is an important application of deformable physics simulation.
Here, a simulated Panda \footnote{FRANKA EMIKA GmbH, Munich, Germany} robot gripper  grasps and deforms a cavity.
For this purpose, we randomly generate cone-shaped cavities with different radii, which are deformed by a gripper from different positions. 
An example simulation step by \model for this task is illustrated in Figure~\ref{fig:figure_one}. 
We use the same amount of samples and data split as in the Tissue Manipulation task.

\section{Results}

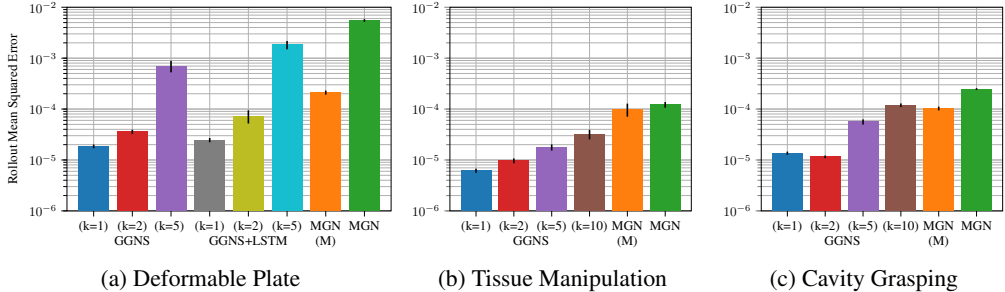
\begin{figure}
     \vspace{-0.5cm}
     \centering 
     \begin{subfigure}{0.38\textwidth} 
         \centering          
         % This file was created with tikzplotlib v0.10.1.
\begin{tikzpicture}[font=\normalsize, scale=0.5]

\definecolor{crimson2143940}{RGB}{214,39,40}
\definecolor{darkgray176}{RGB}{176,176,176}
\definecolor{darkorange25512714}{RGB}{255,127,14}
\definecolor{darkturquoise23190207}{RGB}{23,190,207}
\definecolor{forestgreen4416044}{RGB}{44,160,44}
\definecolor{goldenrod18818934}{RGB}{188,189,34}
\definecolor{gray127}{RGB}{127,127,127}
\definecolor{mediumpurple148103189}{RGB}{148,103,189}
\definecolor{steelblue31119180}{RGB}{31,119,180}

\begin{axis}[
width=10.2cm,
height=7cm,
log basis y={10},
tick align=outside,
tick pos=left,
x grid style={darkgray176},
xmajorgrids,
xmin=-0.69, xmax=7.69,
xtick style={color=black},
xtick={0,1,2,3,4,5,6,7},
xticklabels={
(k=1)\\  ,
(k=2)\\\model,
(k=5)  ,
(k=1)\\  ,
(k=2)\\GGNS+LSTM,
(k=5)\\  ,
MGN\\(M),
MGN
},
xticklabel style={align=center},
y grid style={darkgray176},
ylabel={Rollout Mean Squared Error},
ymajorgrids,
ymin=1e-06, ymax=0.01,
yminorgrids,
ymode=log,
ytick style={color=black}
]
\draw[draw=none,fill=steelblue31119180] (axis cs:-0.4,1e-06) rectangle (axis cs:0.4,1.85631445160619e-05);

\draw[draw=none,fill=crimson2143940] (axis cs:0.6,1e-06) rectangle (axis cs:1.4,3.5661494952661e-05);

\draw[draw=none,fill=mediumpurple148103189] (axis cs:1.6,1e-06) rectangle (axis cs:2.4,0.000705036159797951);

\draw[draw=none,fill=gray127] (axis cs:2.6,1e-06) rectangle (axis cs:3.4,2.46441821257273e-05);

\draw[draw=none,fill=goldenrod18818934] (axis cs:3.6,1e-06) rectangle (axis cs:4.4,7.32757630171599e-05);

\draw[draw=none,fill=darkturquoise23190207] (axis cs:4.6,1e-06) rectangle (axis cs:5.4,0.00183293106644242);

\draw[draw=none,fill=darkorange25512714] (axis cs:5.6,1e-06) rectangle (axis cs:6.4,0.000211342016855876);

\draw[draw=none,fill=forestgreen4416044] (axis cs:6.6,1e-06) rectangle (axis cs:7.4,0.00550452694363064);

\path [draw=black, very thick]
(axis cs:0,1.70270005965326e-05)
--(axis cs:0,2.00992884355911e-05);

\path [draw=black, very thick]
(axis cs:1,3.21349572095228e-05)
--(axis cs:1,3.91880326957993e-05);

\path [draw=black, very thick]
(axis cs:2,0.000525538879921704)
--(axis cs:2,0.000884533439674198);

\path [draw=black, very thick]
(axis cs:3,2.22300018418026e-05)
--(axis cs:3,2.7058362409652e-05);

\path [draw=black, very thick]
(axis cs:4,5.19503387289898e-05)
--(axis cs:4,9.460118730533e-05);

\path [draw=black, very thick]
(axis cs:5,0.00148943497844614)
--(axis cs:5,0.0021764271544387);

\path [draw=black, very thick]
(axis cs:6,0.000192350822438019)
--(axis cs:6,0.000230333211273732);

\path [draw=black, very thick]
(axis cs:7,0.00516381487138)
--(axis cs:7,0.00584523901588129);

\end{axis}

\end{tikzpicture}
         \caption{Deformable Plate}
         \label{fig:trapez_bar_loss_complete_rollout}
     \end{subfigure}% <- nötig
     \begin{subfigure}{.29\textwidth} 
         \centering 
         % This file was created with tikzplotlib v0.10.1.
\begin{tikzpicture}[font=\normalsize, scale=0.5]

\definecolor{crimson2143940}{RGB}{214,39,40}
\definecolor{darkgray176}{RGB}{176,176,176}
\definecolor{darkorange25512714}{RGB}{255,127,14}
\definecolor{forestgreen4416044}{RGB}{44,160,44}
\definecolor{mediumpurple148103189}{RGB}{148,103,189}
\definecolor{sienna1408675}{RGB}{140,86,75}
\definecolor{steelblue31119180}{RGB}{31,119,180}

\begin{axis}[
width=8cm,
height=7cm,
log basis y={10},
tick align=outside,
tick pos=left,
x grid style={darkgray176},
xmajorgrids,
xmin=-0.69, xmax=5.69,
xtick style={color=black},
xtick={0,1,2,3,4,5},
xticklabels={(k=1),
(k=2),
(k=5),
(k=10),
MGN\\(M),
MGN},
xticklabel style={align=center},
y grid style={darkgray176},
ymajorgrids,
ymin=1e-06, ymax=0.01,
yminorgrids,
ymode=log,
extra x ticks ={1.5,2.5},
extra x tick labels={\model},
extra x tick style={
    major tick length=1.4\baselineskip,
    major x tick style={draw=none},
    grid=none,
},
% yticklabels=\empty,
% ytick style={draw=none}
]
\draw[draw=none,fill=steelblue31119180] (axis cs:-0.4,1e-6) rectangle (axis cs:0.4,6.16651565457384e-06);

\draw[draw=none,fill=crimson2143940] (axis cs:0.6,1e-6) rectangle (axis cs:1.4,9.67081102232138e-06);

\draw[draw=none,fill=mediumpurple148103189] (axis cs:1.6,1e-6) rectangle (axis cs:2.4,1.77229315042496e-05);

\draw[draw=none,fill=sienna1408675] (axis cs:2.6,1e-6) rectangle (axis cs:3.4,3.22208209807903e-05);

\draw[draw=none,fill=darkorange25512714] (axis cs:3.6,1e-6) rectangle (axis cs:4.4,9.93794590234756e-05);

\draw[draw=none,fill=forestgreen4416044] (axis cs:4.6,1e-6) rectangle (axis cs:5.4,0.000121950203180313);

\path [draw=black, very thick]
(axis cs:0,5.60496802735819e-06)
--(axis cs:0,6.72806328178949e-06);

\path [draw=black, very thick]
(axis cs:1,8.61545218407963e-06)
--(axis cs:1,1.07261698605631e-05);

\path [draw=black, very thick]
(axis cs:2,1.52622697163969e-05)
--(axis cs:2,2.01835932921023e-05);

\path [draw=black, very thick]
(axis cs:3,2.5370767877407e-05)
--(axis cs:3,3.90708740841736e-05);

\path [draw=black, very thick]
(axis cs:4,7.08833070244187e-05)
--(axis cs:4,0.000127875611022533);

\path [draw=black, very thick]
(axis cs:5,0.000106454943636323)
--(axis cs:5,0.000137445462724304);

\end{axis}

\end{tikzpicture}
         \caption{Tissue Manipulation}
         \label{fig:tissue_bar_loss_complete_rollout}
     \end{subfigure} 
     \begin{subfigure}{.29\textwidth} 
         \centering         
         % This file was created with tikzplotlib v0.10.1.
\begin{tikzpicture}[font=\normalsize, scale=0.5]

\definecolor{crimson2143940}{RGB}{214,39,40}
\definecolor{darkgray176}{RGB}{176,176,176}
\definecolor{darkorange25512714}{RGB}{255,127,14}
\definecolor{forestgreen4416044}{RGB}{44,160,44}
\definecolor{mediumpurple148103189}{RGB}{148,103,189}
\definecolor{sienna1408675}{RGB}{140,86,75}
\definecolor{steelblue31119180}{RGB}{31,119,180}

\begin{axis}[
width=8cm,
height=7cm,
log basis y={10},
tick align=outside,
tick pos=left,
x grid style={darkgray176},
xmajorgrids,
xmin=-0.69, xmax=5.69,
xtick style={color=black},
xtick={0,1,2,3,4,5},
xticklabels={(k=1),
(k=2),
(k=5),
(k=10),
MGN\\(M),
MGN},
xticklabel style={align=center},
y grid style={darkgray176},
ymajorgrids,
ymin=1e-06, ymax=0.01,
yminorgrids,
ymode=log,
extra x ticks ={1.5,2.5},
extra x tick labels={\model},
extra x tick style={
    major tick length=1.4\baselineskip,
    major x tick style={draw=none},
    grid=none,
},
% yticklabels=\empty,
% ytick style={draw=none}
]
\draw[draw=none,fill=steelblue31119180] (axis cs:-0.4,1e-6) rectangle (axis cs:0.4,1.36128582060337e-05);

\draw[draw=none,fill=crimson2143940] (axis cs:0.6,1e-6) rectangle (axis cs:1.4,1.15319627026717e-05);

\draw[draw=none,fill=mediumpurple148103189] (axis cs:1.6,1e-6) rectangle (axis cs:2.4,5.61035081744194e-05);

\draw[draw=none,fill=sienna1408675] (axis cs:2.6,1e-6) rectangle (axis cs:3.4,0.000119192347703774);

\draw[draw=none,fill=darkorange25512714] (axis cs:3.6,1e-6) rectangle (axis cs:4.4,0.000103072641293208);

\draw[draw=none,fill=forestgreen4416044] (axis cs:4.6,1e-6) rectangle (axis cs:5.4,0.000246960852543513);

\path [draw=black, very thick]
(axis cs:0,1.2621197010713e-05)
--(axis cs:0,1.46045194013544e-05);

\path [draw=black, very thick]
(axis cs:1,1.08419677329353e-05)
--(axis cs:1,1.22219576724081e-05);

\path [draw=black, very thick]
(axis cs:2,4.9541341389836e-05)
--(axis cs:2,6.26656749590027e-05);

\path [draw=black, very thick]
(axis cs:3,0.000109517221598376)
--(axis cs:3,0.000128867473809172);

\path [draw=black, very thick]
(axis cs:4,9.43293592293738e-05)
--(axis cs:4,0.000111815923357042);

\path [draw=black, very thick]
(axis cs:5,0.000236388867833816)
--(axis cs:5,0.000257532837253209);

\end{axis}

\end{tikzpicture}
         \caption{Cavity Grasping}
         \label{fig:tube_bar_loss_complete_rollout}
     \end{subfigure} 
     \caption{Rollout Mean Squared Error of \model and \gls{mgn} baselines evaluated on the test set of the three datasets. We report the results for \model using point clouds in every $k$-th time step. \gls{mgn}(M) indicates the baseline method of \gls{mgn} that uses the ground truth material as input feature. \model outperforms the \gls{mgn} baseline in all settings and in most cases even if it has access to the complete initial state. For the Deformable Plate task we additionally report the errors for GGNS+LSTM, which perform worse than \model for all $k$ but still outperforming the \gls{mgn} baseline.} 
     \label{fig:complete_rollout}
     \vspace{-0.25cm}
\end{figure} 

\begin{figure}
     \centering 
     \begin{subfigure}{.33\textwidth} 
         \centering 
         % This file was created with tikzplotlib v0.10.1.
\begin{tikzpicture}[font=\normalsize, scale=0.5]

\definecolor{crimson2143940}{RGB}{214,39,40}
\definecolor{darkgray176}{RGB}{176,176,176}
\definecolor{darkorange25512714}{RGB}{255,127,14}
\definecolor{forestgreen4416044}{RGB}{44,160,44}
\definecolor{mediumpurple148103189}{RGB}{148,103,189}
\definecolor{sienna1408675}{RGB}{140,86,75}
\definecolor{steelblue31119180}{RGB}{31,119,180}

\begin{axis}[
width=8cm,
tick align=outside,
tick pos=left,
x grid style={darkgray176},
xmajorgrids,
xmin=-0.69, xmax=4.69,
xtick style={color=black},
xtick={0,1,2,3,4},
xticklabels={(k=1)\\  ,
(k=2)\\\model,
(k=5)\\  ,
MGN\\(M),
MGN},
xticklabel style={align=center},
y grid style={darkgray176},
ylabel={50-step Rollout IoU},
ymajorgrids,
ymin=0.8, ymax=1,
ytick style={color=black}
]
\draw[draw=none,fill=steelblue31119180] (axis cs:-0.4,0) rectangle (axis cs:0.4,0.99704621025959);

\draw[draw=none,fill=crimson2143940] (axis cs:0.6,0) rectangle (axis cs:1.4,0.995112733983007);

\draw[draw=none,fill=mediumpurple148103189] (axis cs:1.6,0) rectangle (axis cs:2.4,0.975341214264558);

\draw[draw=none,fill=darkorange25512714] (axis cs:2.6,0) rectangle (axis cs:3.4,0.98516182808192);

\draw[draw=none,fill=forestgreen4416044] (axis cs:3.6,0) rectangle (axis cs:4.4,0.908239568357729);

\path [draw=black, very thick]
(axis cs:0,0.996956365010978)
--(axis cs:0,0.997136055508202);

\path [draw=black, very thick]
(axis cs:1,0.994900672689944)
--(axis cs:1,0.99532479527607);

\path [draw=black, very thick]
(axis cs:2,0.970345580204526)
--(axis cs:2,0.98033684832459);

\path [draw=black, very thick]
(axis cs:3,0.984472877374851)
--(axis cs:3,0.985850778788989);

\path [draw=black, very thick]
(axis cs:4,0.902493106974874)
--(axis cs:4,0.913986029740585);

\end{axis}

\end{tikzpicture}
            \caption{Deformable Plate IoU}
            \label{fig:trapez_bar_IOU_complete_rollout}
     \end{subfigure}% 
     \begin{subfigure}{.33\textwidth} 
         \centering 
         % This file was created with tikzplotlib v0.10.1.
\begin{tikzpicture}[font=\normalsize, scale=0.5]
 %\footnotesize

\definecolor{crimson2143940}{RGB}{214,39,40}
\definecolor{darkgray176}{RGB}{176,176,176}
\definecolor{darkorange25512714}{RGB}{255,127,14}
\definecolor{forestgreen4416044}{RGB}{44,160,44}
\definecolor{mediumpurple148103189}{RGB}{148,103,189}
\definecolor{orchid227119194}{RGB}{227,119,194}
\definecolor{steelblue31119180}{RGB}{31,119,180}

\begin{axis}[
width=8cm,
tick align=outside,
tick pos=left,
x grid style={darkgray176},
xmajorgrids,
xmin=-0.69, xmax=5.69,
xtick style={color=black},
xtick={0,1,2,3,4,5},
xticklabels={(k=1)\\  ,
(k=2)\\\model,
(k=5)\\  ,
MGN\\(M),
MGN,
Alpha\\Shapes},
xticklabel style={align=center},
y grid style={darkgray176},
ylabel={50-step Rollout IoU},
ymajorgrids,
ymin=0.8, ymax=1,
ytick style={color=black}
]
\draw[draw=none,fill=steelblue31119180] (axis cs:-0.4,0) rectangle (axis cs:0.4,0.961054071297998);

\draw[draw=none,fill=crimson2143940] (axis cs:0.6,0) rectangle (axis cs:1.4,0.927740405865235);

\draw[draw=none,fill=mediumpurple148103189] (axis cs:1.6,0) rectangle (axis cs:2.4,0.887938378604596);

\draw[draw=none,fill=darkorange25512714] (axis cs:2.6,0) rectangle (axis cs:3.4,0.888197856175468);

\draw[draw=none,fill=forestgreen4416044] (axis cs:3.6,0) rectangle (axis cs:4.4,0.853708858482565);

\draw[draw=none,fill=orchid227119194] (axis cs:4.6,0) rectangle (axis cs:5.4,0.880968238155381);

\path [draw=black, very thick]
(axis cs:0,0.953285862040423)
--(axis cs:0,0.968822280555573);

\path [draw=black, very thick]
(axis cs:1,0.916408397217435)
--(axis cs:1,0.939072414513036);

\path [draw=black, very thick]
(axis cs:2,0.87261854655937)
--(axis cs:2,0.903258210649823);

\path [draw=black, very thick]
(axis cs:3,0.871259556703086)
--(axis cs:3,0.905136155647849);

\path [draw=black, very thick]
(axis cs:4,0.843154714822464)
--(axis cs:4,0.864263002142666);

\path [draw=black, very thick]
(axis cs:5,0.880968238155381)
--(axis cs:5,0.880968238155381);
%0.9483711990663698 (with 800 nodes)

\end{axis}

\end{tikzpicture}
            \caption{IoU without initial mesh}
            \label{fig:automesh_bar_IoU_complete_rollout}
     \end{subfigure}% 
     \begin{subfigure}{.33\textwidth} 
         \centering         
         % This file was created with tikzplotlib v0.10.1.
\begin{tikzpicture}[font=\normalsize, scale=0.5]

\definecolor{darkorange25512714}{RGB}{255,127,14}
\definecolor{darkslategray38}{RGB}{38,38,38}
\definecolor{forestgreen4416044}{RGB}{44,160,44}
\definecolor{lightgray204}{RGB}{204,204,204}
\definecolor{steelblue31119180}{RGB}{31,119,180}
\definecolor{darkgray176}{RGB}{176,176,176}

\begin{axis}[
width=8cm,
axis line style={black},
legend cell align={left},
legend style={
  fill opacity=0.9,
  draw opacity=1,
  text opacity=1,
  at={(0.03,0.03)},
  anchor=south west,
  draw=darkgray176
},
tick align=outside,
tick pos=left,
x grid style={darkgray176},
xlabel=\textcolor{black}{k},
xmajorgrids,
xmin=1, xmax=10,
xtick style={color=black},
xtick={1,2,3,4,5,6,7,8,9,10},
y grid style={darkgray176},
ylabel=\textcolor{black}{Normalized Benefit},
ymajorgrids,
ymin=0, ymax=1.2,
ytick style={color=black}
]
\path [fill=steelblue31119180, fill opacity=0.25]
(axis cs:1,0.99972260334927)
--(axis cs:1,1.00027739665073)
--(axis cs:2,0.99766221167003)
--(axis cs:3,0.995565903991501)
--(axis cs:4,0.962861978923999)
--(axis cs:5,0.906999650581131)
--(axis cs:6,0.829352582949572)
--(axis cs:7,0.718384052400359)
--(axis cs:8,0.577505230400481)
--(axis cs:9,0.467312138371966)
--(axis cs:10,0.389505568589928)
--(axis cs:10,0.178814922527434)
--(axis cs:10,0.178814922527434)
--(axis cs:9,0.283294068298324)
--(axis cs:8,0.390663613018257)
--(axis cs:7,0.515056649418193)
--(axis cs:6,0.651285609703674)
--(axis cs:5,0.843675782485073)
--(axis cs:4,0.863393666798018)
--(axis cs:3,0.959486918828108)
--(axis cs:2,0.995517449160899)
--(axis cs:1,0.99972260334927)
--cycle;

\path [fill=darkorange25512714, fill opacity=0.25]
(axis cs:1,0.989140031328542)
--(axis cs:1,1.01085996867146)
--(axis cs:2,0.953881667844091)
--(axis cs:3,0.899958931441196)
--(axis cs:4,0.853910994220759)
--(axis cs:5,0.828338631860563)
--(axis cs:6,0.788800740636174)
--(axis cs:7,0.742428162095171)
--(axis cs:8,0.700355285789124)
--(axis cs:9,0.66419342078132)
--(axis cs:10,0.635523517232466)
--(axis cs:10,0.375459616988259)
--(axis cs:10,0.375459616988259)
--(axis cs:9,0.423730490453277)
--(axis cs:8,0.477020424135098)
--(axis cs:7,0.539897204816255)
--(axis cs:6,0.649950742770284)
--(axis cs:5,0.735791621106229)
--(axis cs:4,0.778033535503365)
--(axis cs:3,0.849735941149463)
--(axis cs:2,0.913500566602662)
--(axis cs:1,0.989140031328542)
--cycle;

\path [fill=forestgreen4416044, fill opacity=0.25]
(axis cs:1,0.995852782458458)
--(axis cs:1,1.00414721754154)
--(axis cs:2,1.01195688999745)
--(axis cs:3,0.986706818524126)
--(axis cs:4,0.938323642073783)
--(axis cs:5,0.846472430453523)
--(axis cs:6,0.760482520265279)
--(axis cs:7,0.703897310140814)
--(axis cs:8,0.660382679191639)
--(axis cs:9,0.620600942440337)
--(axis cs:10,0.58901019348923)
--(axis cs:10,0.50608480496542)
--(axis cs:10,0.50608480496542)
--(axis cs:9,0.552215110077635)
--(axis cs:8,0.594492469021965)
--(axis cs:7,0.642814495968193)
--(axis cs:6,0.703677141356201)
--(axis cs:5,0.790744381612997)
--(axis cs:4,0.89091895741777)
--(axis cs:3,0.971910730731762)
--(axis cs:2,1.00616023483282)
--(axis cs:1,0.995852782458458)
--cycle;

\addplot [line width=1.4pt, forestgreen4416044]
table {%
1 1
2 1.00905856241513
3 0.979308774627944
4 0.914621299745777
5 0.81860840603326
6 0.73207983081074
7 0.673355903054504
8 0.627437574106802
9 0.586408026258986
10 0.547547499227325
};
\addlegendentry{Cavity}
\addplot [line width=1.4pt, darkorange25512714]
table {%
1 1
2 0.933691117223376
3 0.87484743629533
4 0.815972264862062
5 0.782065126483396
6 0.719375741703229
7 0.641162683455713
8 0.588687854962111
9 0.543961955617299
10 0.505491567110362
};
\addlegendentry{Tissue}
\addplot [line width=1.4pt, steelblue31119180]
table {%
1 1
2 0.996589830415465
3 0.977526411409804
4 0.913127822861008
5 0.875337716533102
6 0.740319096326623
7 0.616720350909276
8 0.484084421709369
9 0.375303103335145
10 0.284160245558681
};
\addlegendentry{Plate}
\end{axis}

\end{tikzpicture}
            \caption{Different grounding frequencies} 
            \label{fig:k_hop}
     \end{subfigure}% 
     \caption{(\subref{fig:trapez_bar_IOU_complete_rollout}) Rollout \gls{iou} for the Deformable Plate dataset for \model and the \gls{mgn} baseline. The main findings here are similiar to the MSE results but enable to compare against the values in (\subref{fig:automesh_bar_IoU_complete_rollout}), where the initial mesh is created from initial the point cloud. The achieved \glspl{iou} are lower compared to using the ground truth mesh, but \model ($k\leq 2$) still outperforms all baselines including the Alpha Shapes in this setting.
     (\subref{fig:k_hop}) Normalized benefit of using a point cloud in every $k$-th timestep, where $k=1$ means a point cloud is available in every time step.} 
     \label{fig:additional_results}
     \vspace{-0.25cm}
\end{figure}
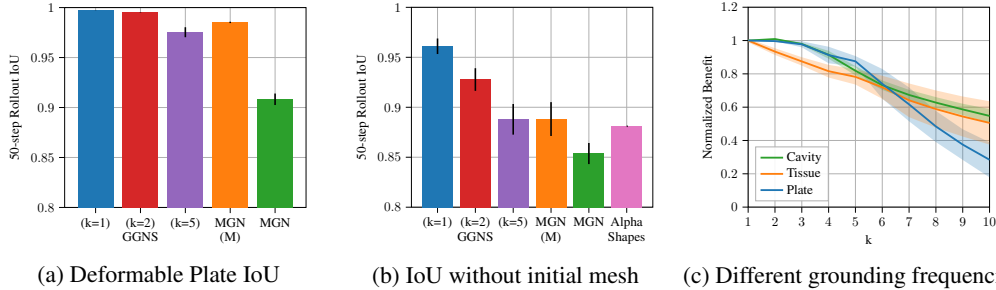

\textbf{Main Results.}
We test our method on the three deformation prediction tasks described in Section~\ref{sec:experiments} and compare it to \gls{mgn} with and without material information.
We find that \model can use the point cloud information to produce high quality rollouts that closely match the true system states. 
An example is shown in Figure~\ref{fig:trapez_qualitative}, which visualizes the final simulated meshes for our method and the ground truth simulation. 
Additionally, \model outperforms the baselines even when they have access to the complete initial state, which our model has not. 
Figure~\ref{fig:tissue_qualitative} shows the qualitative differences between \model and \gls{mgn} on the Tissue Manipulation task.
Additional visualizations for all tasks and both methods can be found in Appendix~\ref{app_sec:qualitative_results}. 
The evaluations for full rollouts are given in Figure~\ref{fig:complete_rollout}.
Table~\ref{tab:trapezmpc} shows results for the $m+10$-step evaluation.
Appendix~\ref{app_sec:ablations} shows the performance of \gls{ggns} for different model hyperparameters.
% , as well as for noisy point clouds and scenes with fewer cameras and resulting partial occlusions.
Similar to~\citet{pfaff2020learning}, we find that \model is robust to most parameter choices, and that a modest amount of training noise is crucial for long-term rollouts.
To show the applicability of our method for more realistic point cloud data, we provide additional ablations on noisy and partial observable point clouds in Appendix~\ref{app_sec:ablations}. 
We find that our model is quite robust to the quality of the point clouds and can still reliably use their information to ground the simulation. 
On the Deformable Plate dataset, we additionally evaluate the mean \glsfirst{iou} during the rollouts to emphasize the compliance with the overall shape of the object rather than that of individual mesh nodes.
%, highlighting, for example, the simulation of the correct material.
The results are illustrated in Figure~\ref{fig:trapez_bar_IOU_complete_rollout}.

\textbf{Recurrent Imputation Model.}
For the $2d$ data of the Deformable Plate task, we additionally compare our imputation model to the GGNS+LSTM approach, which can use the recurrence of LSTMs to pass information over time. Figure~\ref{fig:trapez_bar_loss_complete_rollout} shows that \model outperforms this alternative approach for each $k$.
We find that our simple architecture outperforms the recurrent one while requiring significantly less time to train, likely due to the additional complexity of training the recurrent model. The qualitative results in Appendix~\ref{app_sec:qualitative_results} confirm these findings.
% The current limitations of the method therefore likely lie more in the focus on $1$-step predictions rather than in the lack of recurrence. For this reason, we use the simpler imputation method presented here, which is accompanied by a significant reduction in computation time. 

\textbf{Initial Mesh Generation.} 
Using the \gls{iou} metric, we can compare objects across different mesh representations. 
The results in Figure~\ref{fig:automesh_bar_IoU_complete_rollout} show that \model produces accurate rollouts even if the initial mesh is generated directly from the initial point cloud. 
For this, we compute a mesh with similar resolution to the training meshes from the convex hull of the initial point cloud, avoiding the dependence on \textit{any} simulation data.
This procedure marks an important step towards using these models on real world data. 
The results indicate that generating the initial mesh from point cloud information results in a degradation of the performance compared to an evaluation that uses a provided mesh.
Yet, it still allows for a high-quality prediction of the deformation. 
%to using ground truth meshes, but comes with the advantage that no such mesh is required during evaluation.
%This is particularly interesting for the use in a robotic system, because no initial mesh of the deformable object to be simulated is required. 
The comparison to Alpha Shapes shows that combining infrequent point cloud information ($k=5$) with a simulator leads to better and more consistent results than directly creating the mesh from the point cloud in each time step.
Additionally, our model naturally tracks the correspondences of mesh nodes  over time, whereas Alpha Shapes cannot observe the evolution of individual particles in the system. As such, \gls{ggns} allows for a more thorough understanding of the modeled process.
% \todo{maybe add a qualitative result to showcase consistency in appendix}
% Overall, the \gls{mgn} baseline method is still outperformed in this case, for $k\leq2$ even if it can rely on the ground truth mesh or the material information.  

\textbf{Grounding Frequency.} 
Figure~\ref{fig:k_hop} shows the normalized performance of \model for grounding frequencies $k \in \{1..10\}$ across tasks. Here, a value of $1.0$ corresponds to the performance for $k=1$, and $0.0$ to the performance \gls{mgn}.
% Therefore, the values are normalized to the performance of \model with $k=1$ being $1.0$ and the performance of \gls{mgn} being $0.0$. 
For all tasks there is a clear advantage in utilizing the point cloud information, and the performance increases with the frequency of available point clouds. 
% Figure~\ref{fig:k_hop} shows this relation for $k \in \{1..10\}$. 
% Therefore, the values are normalized to the performance of \model with $k=1$ being $1.0$ and the performance of \gls{mgn} being $0.0$. 
% On the cavity deformation task, the best performance is actually obtained for $k=2$ which is likely due to regularization artifact caused by the training method. 

% \todo{add ablation plot!}
% \textbf{Ablations.}
% \textit{Hyperparameters.}
% \begin{itemize}
%     \item General hyperparameters of the network
%     \item Noise level
%     \item Radius of the observation graph
% \end{itemize}

\begin{table}
    \vspace{-0.5cm}
	\begin{center}
		\begin{tabular}{lccc}
			\toprule
			\multirow{2}{*}{Approach} & \multicolumn{3}{c}{$m+10$-step MSE $\times 10^{-5}$} \\
            & Plate & Tissue & Cavity \\
			\midrule
			\model & $2.907 \pm 0.172$ & $0.514 \pm 0.052$ &  $0.923 \pm 0.040$ \\
			MGN (M) & $10.663 \pm 1.063$ & $5.027 \pm 1.489$  & $6.294 \pm 0.668$  \\
			MGN & $282.684 \pm 18.112$ & $5.885 \pm 0.723$ & $11.528 \pm 0.747$  \\
			\bottomrule
		\end{tabular}
	\end{center}
    \caption{Evaluation on the $m+10$-step prediction setting on the test set for all three tasks. \model clearly outperforms the baselines on all tasks even if they have access to the full initial simulation state.}
	\label{tab:trapezmpc}
\end{table}

% \begin{figure}
%      \centering 
%      \begin{subfigure}{0.3\textwidth} 
%          \centering 
%          \label{fig:trapez_bar_loss_complete_rollout}
%          \input{tikz/results/trapez_bar_loss_complete_rollout.tex}
%          \caption{Deformable Plate}
%      \end{subfigure}% <- nötig
%      \begin{subfigure}{.33\textwidth} 
%          \centering 
%          \label{fig:tissue_bar_loss_complete_rollout}
%          \input{tikz/results/tissue_bar_loss_complete_rollout.tex}
%          \caption{Tissue Manipulation}
%      \end{subfigure} 
%      \begin{subfigure}{.33\textwidth} 
%          \centering 
%          \label{fig:tube_bar_loss_complete_rollout}
%          \input{tikz/results/tube_bar_loss_complete_rollout.tex}
%          \caption{Cavity Grasping}
%      \end{subfigure} 
%      \caption{Rollout Mean Squared Error of \model and \gls{mgn} baselines evaluated on the test set of the three tasks. We report the results for \model using point clouds in every $k$-th time step. \gls{mgn} (M) indicates the baseline method of \gls{mgn} that uses the ground truth material as input feature. \model outperforms the \gls{mgn} baseline in all settings and in most cases even if it has access to the complete initial state.} 
% \end{figure} 

\section{conclusion}
We propose \glsfirst{ggns}, an extension of the popular Graph Network Simulator framework that can utilize auxiliary observations to accurately simulate complex dynamics from incomplete initial system states.
Utilizing a neighborhood graph computed from point cloud information and an imputation-based training scheme, our model is able to \textit{ground} its prediction in an observation of the true system state.
We show experimentally that this leads to high-quality simulations in challenging $2d$ and $3d$ object deformation tasks, outperforming existing approaches even when these are provided with full information about the system.

In future work, we will extend \glspl{ggns} to explicitly model uncertainty and maintain a belief over the latent variables of the system, e.g., by employing a Kalman filter in a learned latent space \citep{becker2019recurrent}. 
Another promising direction is to adapt the current next-step prediction loss to instead predict a trajectory over a small period of time to increase the long-term consistency of the model. Finally, we will employ our model for model-predictive control and model-based Reinforcement Learning in both simulation and on a real robot.

% \begin{enumerate}
%     \item Repeat and rephrase the introduction.
%     \item Mention the most significant detail and contribution of our method.
%     \item (Maybe) explain some limitations that we have.
%     \item For future work, we want to 
%     \begin{enumerate}
%         \item extend the model to an explicit consideration of uncertainty, e.g., by combining it with recent networks employing Kalman Smoothing (RKN). This would come with the additional benefit of presenting a natural framework for sensor fusion, allowing us to input observation modalities of different frequencies in a principled manner.
%         \item improve upon the current $1$-step prediction loss by instead predicting for each step the movement of each mesh node over a small period of time (using e.g., movement primitives, though we may not want to say that as it would need more explanation)
%         \item use the model for MPC experiments, including ones on a real robot
%     \end{enumerate}
%     \item we could also consider a limitations section in which we state what currently does not work and why. This is usually pretty need and useful, but some reviewers may take it as an easy way to find flaws in the current approach. 
% \end{enumerate}


\subsubsection*{Acknowledgments}
We thank Vincent Kreuziger for the helpful discussions on the visualizations and for the high-quality blender renderings. 
The authors acknowledge support by the state of Baden-Württemberg through bwHPC.
GN was supported by the DFG research unit DFG-FOR 5339 (AI-based Methodology for the Fast Maturation 
of Immature Manufacturing Processes) and GN and NF were supported by the BMBF project Davis 
(Datengetriebene Vernetzung für die ingenieurtechnische Simulation).
% % todo mention vincent :)
% antything else?
\bibliography{bibliography}
\bibliographystyle{iclr2023_conference}

\clearpage
\appendix
\section{Model Details}
\label{app_sec:detailed_model}

\begin{figure}
    \centering
	\includegraphics[width=0.9\textwidth]{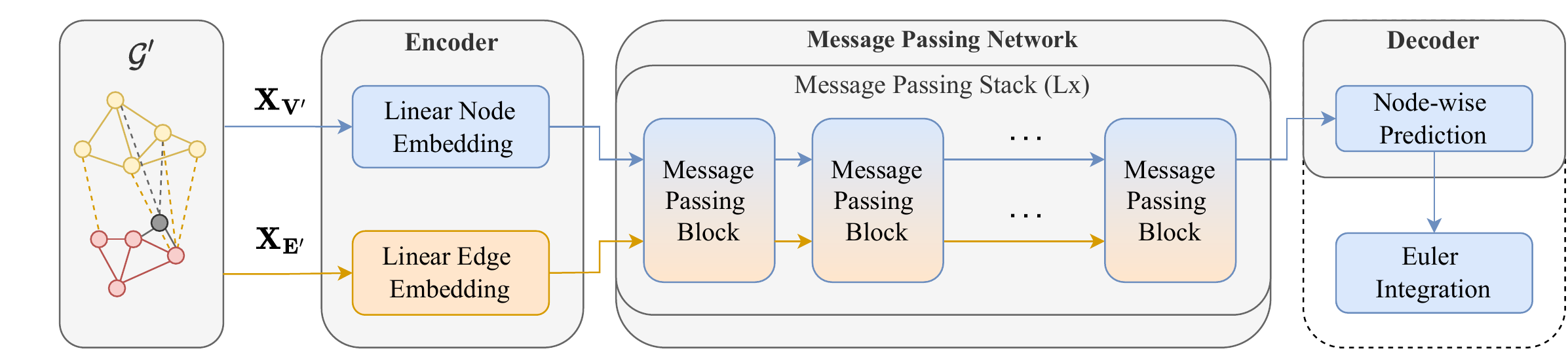}
	\caption{A detailed view of the \gls{gnn} part of \model. Given a graph $\gG'$, the node and edge features $\mathbf{X}_{\mathbf{V}'}$ and $\mathbf{X}_{\mathbf{E}'}$ are linearly embedded into a latent space and then updated with $L$ Message Passing Blocks. The resulting predictions are interpreted as dynamic quantities that are used to update the system.}
    \label{fig:detailed_model}
    \vspace{0.25cm}
\end{figure}

The Message Passing Network employed by \model is displayed in \ref{fig:detailed_model}. As node-wise predictions we use velocities, which are Euler-integrated once to update the positions of the mesh of the deformable object. 

\section{Environment Details}
\label{app_sec:environments}

%Detailed listing of all environments. Also, additional details on how the meshes and pointclouds etc. are constructedMaybe reference to last part in the main paper
Here, we describe all key aspects, which are valid for all three environments. 
All datasets are simulated using \gls{sofa} and include different material properties. 
Therefore, we choose discrete Poisson's ratios from $\nu \in \{-0.9, 0.0, 0.49\}$ for one-third of all simulated trajectories each. 
Other material parameters are kept constant, e.g., for the mass we choose large values for the solid object and smaller values for the deformable to ensure sufficient deformation.
The chosen parameters do not represent the full reality, as there are other material parameters that could be varied.
However, as we want to showcase the capabilities of our method, we selected these parameters as they displayed the biggest impact on the deformation behavior.

\subsection{Point Cloud Generation}
The required point clouds are not directly available in \gls{sofa}, but instead rendered from the scene of the meshes using \textit{Raycasting} from Open3D \citep{zhou2018open3d}.
We therefore place virtual cameras around and on top of the scene to generate partial point clouds from different directions. For the Deformable Plate dataset one camera is sufficient, while the other two tasks rely on four cameras around and one camera on top of the scene.
This results in a good, but not complete coverage of the entire surface with points of the point cloud.
Even though there are five cameras around the scene, there are areas that are not covered: For the tissue, the parts that are occluded by the red liver, and for the cavity, parts of the inner surface depending on how the upper and lower radii deviates from one another. 
Also, as there can be no camera from below, there are naturally no points on the lower surface for both datasets.
In Appendix~\ref{app_sec:ablations} we additionally provide results for less cameras on the cavity dataset, leading to only partially observable point clouds.
If more than one point cloud camera is used, the resulting point clouds are fused and subsampled accordingly to achieve a processable number of points. 
We voxel subsample in world space, so the points do not belong to any specific part of the mesh, but can rather be seen as some “interpolation” between mesh vertexes. 
The main challenge is that there are no point correspondences and that the model needs to figure out which point of the point cloud belongs to which vertex in the mesh to do the correction of the mesh nodes for grounding the simulation.
Still, voxel subsampling leads to the most structured results compared to other subsampling techniques, which helps the model to account for correspondences between points over time. \\ \\

\subsection{Input Features}
In addition to encoding the node or edge type as one-hot features, we add an encoding to static nodes and encode the velocity of the collider in its node features. 
We encode the positions in space as relative features in the edges instead of absolute encodings in the node features following previous work \citep{sanchezgonzalez2020learning}. All edges thus receive their relative world coordinates, while mesh edges additionally contain relative coordinates in mesh space.

% \todo{For each environment, briefly describe \textit{why} we choose it and what makes it interesting.}
% \todo{Go into more detail for each environment and its intricacies (which parameters, how many points of data and so on. Most of this will go into the appendix eventually, but thats fine}

\subsection{Collision Handling}
\gls{sofa} as the ground truth simulator handles collision between objects using triangular surface meshes of all objects involved to detect collisions. 
The detection is implemented using the \textit{LocalMinDistance} method and detected collisions are included in the constraints of the system. 
Using Lagrangian multipliers, the constraints are then processed together with the other forces from the deformation to solve the complete FEM system \citep{faure2012sofa}. 
In contrast to that, \model uses one-hot encoded edges between the rigid and the soft body that are used by the model to compute the dynamics. There is no explicit handling of collisions, the network learns to avoid them and adapts the mesh accordingly.

\subsection{Deformable Plate}
For this environment, we simulate a family of $2$-dimensional trapezoids deformed by a circular collider with constant velocity. 
We vary the size of the collider by sampling from a triangular distribution between 15 and $60~\si{\percent}$ of the edge length of the deformable object. 
For the collider start position we sample from a uniform distribution between the left and right corner of deformable object.
We record $50$ time steps per trajectory and $945$ trajectories in total, which are split in $675/135/135$ trajectories per train, evaluation and test set. 
A single data sample contains approx. $700$ nodes: $57$ nodes for the collider, $81$ nodes for the mesh oft the deformable object and around $600$ points in the subsampled point cloud. 
The mesh itself consists of $416$ edges, the total number of edges is about $3$ K depending on the deformation in the according time step.
In contrast to the Poisson's ratio, the other adjustable material parameter in SOFA, the Young's modulus is kept constant for all samples at $E = 5\,000 \si{\Pa}$. 
It describes the compressive stiffness when a force is applied lengthwise.
The different material properties together with the different trapezoidal shapes introduce uncertainty in the form of multi-modality into the data. 
The reason for this is that different deformations result in states that cannot be clearly assigned to a single trapez-material combination. 
We construct this dataset because it comes with lower computational cost due to the restriction to $2$d, but already allows for more general statements due to the non-trivial deformations and the multi-modality. 
Therefore, it is especially suitable as a proof-of-concept and for ablations.

%$1$ K to $4$ K edges depending on the graph connectivity setting; here: $3$K 
%\todo{Mention different materials and initial states and resulting uncertainty/multi-modality.}

\subsection{Tissue Manipulation}
Here, a piece of tissue is deformed by a rigid gripper which could be part of a robot-assisted surgery scenario.
To generate diversity, we generate random motions in a $2d$ plane and sample a random gripping point from the $19$ top mesh points. 
We record $100$ time steps per trajectory and $840$ trajectories in total, which we split in $600/120/120$ trajectories per train, evaluation and test set. 
A single data sample consists of approx. $1\,200$ nodes: $361$ for the mesh, one for the gripper and about $850$ for the point cloud. The mesh consists of $2\,154$ edges, which leads to a total number of about $3\,800$ edges depending on the time step. 
To ensure physically plausible deformation, each Poisson's ratio is assigned its specific Young's modulus from $E \in \{10\,000, 80\,000, 30\,000\} \si{\Pa}$.
If instead it were kept the same for each Poisson's ratio, the gripper could penetrate the deformable object or pull it without touching it.
The uncertainty in this dataset is mainly in the initial state, which can result in different deformations depending on the material from the same initial state.

% \todo{...}.
% \todo{
% Mention and highlight shift problem with MGN, and how this is much better with our approach -> in Results?}
% ranges between $1.8$ K to $9$ K depending on the graph connectivity setting chosen; here: $3.8$ K.

\subsection{Cavity Grasping}
We randomly generate cone-shaped cavities with radii between $87.5 \si{\percent}$ and $50 \si{\percent}$ of the maximum possible gripping width. 
The cone shape helps to increase uncertainty in the form of multi-modality in the data, because the states resulting from deformation cannot be clearly assigned to a single cone-material combination. 
The deformable cavities are deformed by a gripper located at random positions in space. 
The positions are sampled form a hexahedron around the geometrical center of the cavity ensuring collision free starting positions. 
For the grasping, the gripper moves as quickly as it is allowed to the gripping position and then closes its fingers with constant velocity. 
We record $100$ time steps per trajectory and $840$ trajectories in total, which are split in $600/120/120$ trajectories per train, evaluation and test set. 
A single data sample consists of approx. $2.4$ K nodes: $750$ for the mesh, $636$ for the gripper and about $1$ K for the point cloud. 
The mesh consists of $4\,500$ edges, the overall number of edges in the graph is about $8.5$ K depending on the exact time step.
The motivation for the creation of this environment is that a successful use of our method in this setting is an important step on the way to a real-world application.  
%\todo{The $3d$ cavity dataset we discussed. Most complex task, (hopefully) interesting and promising results}

%  ranges between $5$ K to $10$ K depending on the graph connectivity setting chosen; here: $8.5$ K.

%\label{ssec:experiments:ablations}
\clearpage
\section{Qualitative Results}
\label{app_sec:qualitative_results}
\begin{figure}
    \centering
	\includegraphics[width=0.82\textwidth]{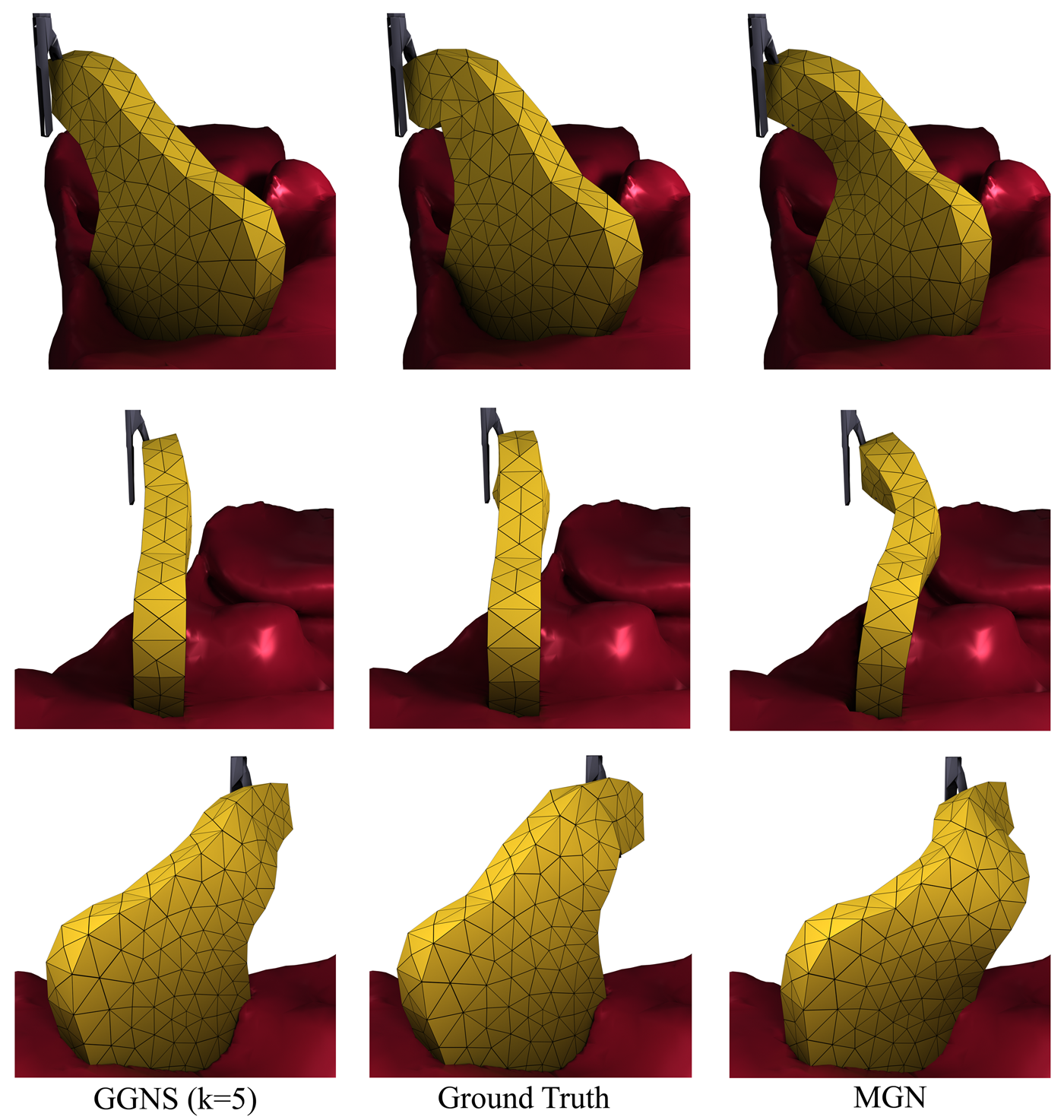}
	\caption{Visualization of a test trajectory in the Tissue Manipulation dataset from three different viewing angles (rows) at time step $t=70$ for \model (left), the ground truth simulation (middle) and \gls{mgn} (right).}
    \label{app_fig:tissue_qualitative_left}
\end{figure}

\begin{figure}
    \centering
	\includegraphics[width=0.70\textwidth]{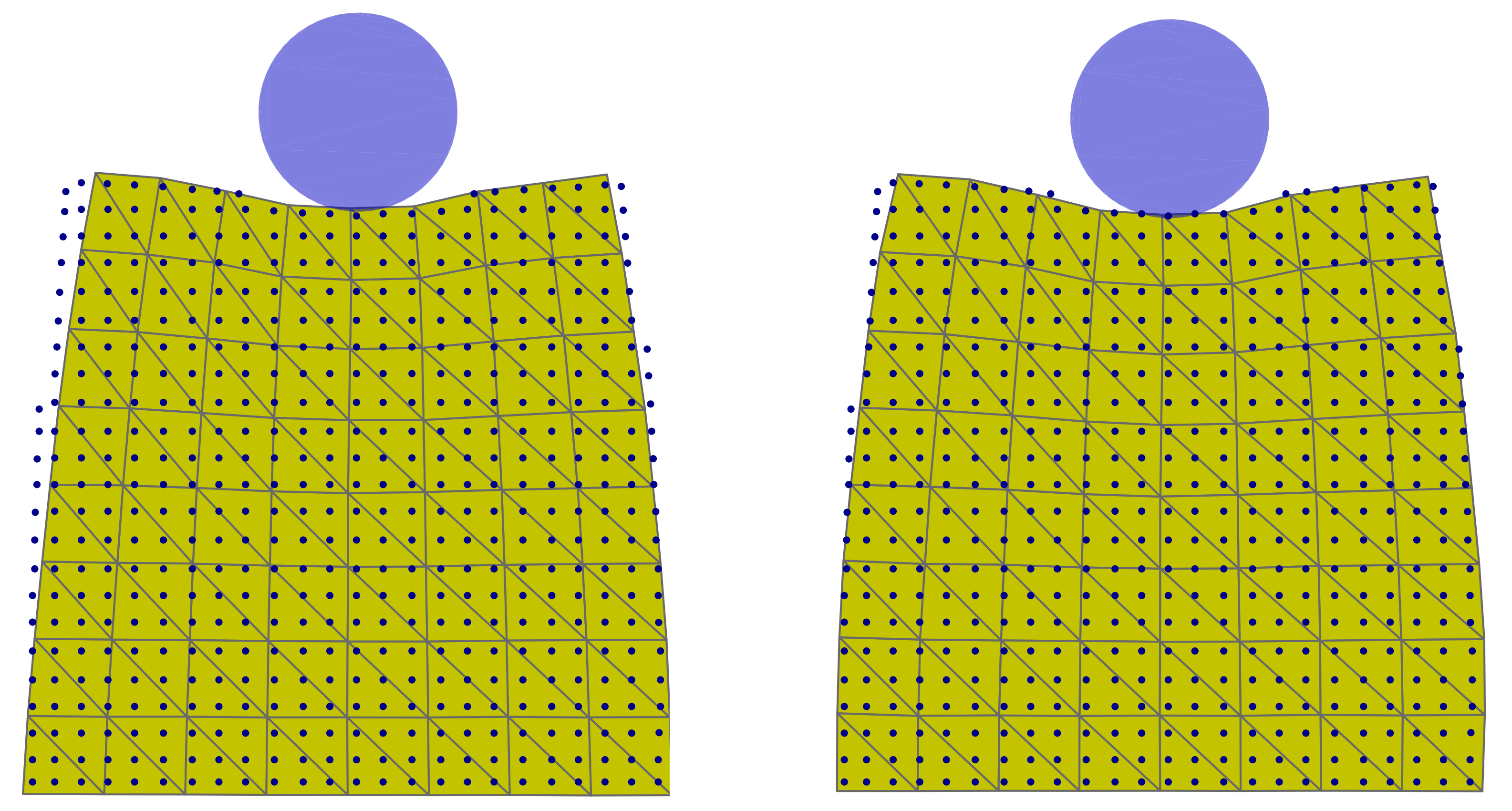}
	\caption{Overlay of the point cloud and the predicted mesh for two consecutive time steps $t=[10,11]$ in the Deformable Plate dataset. We repeat the point cloud from the earlier simulation step in both images for clarity. The illustration shows the correction behavior of \model by including the point cloud to ground the mesh based simulation in this time step. This can be observed particularly well in the upper left and right corners of the plate.}
    \label{app_fig:figure1_plate}
\end{figure}

\begin{figure}
    \centering
	\includegraphics[width=0.70\textwidth]{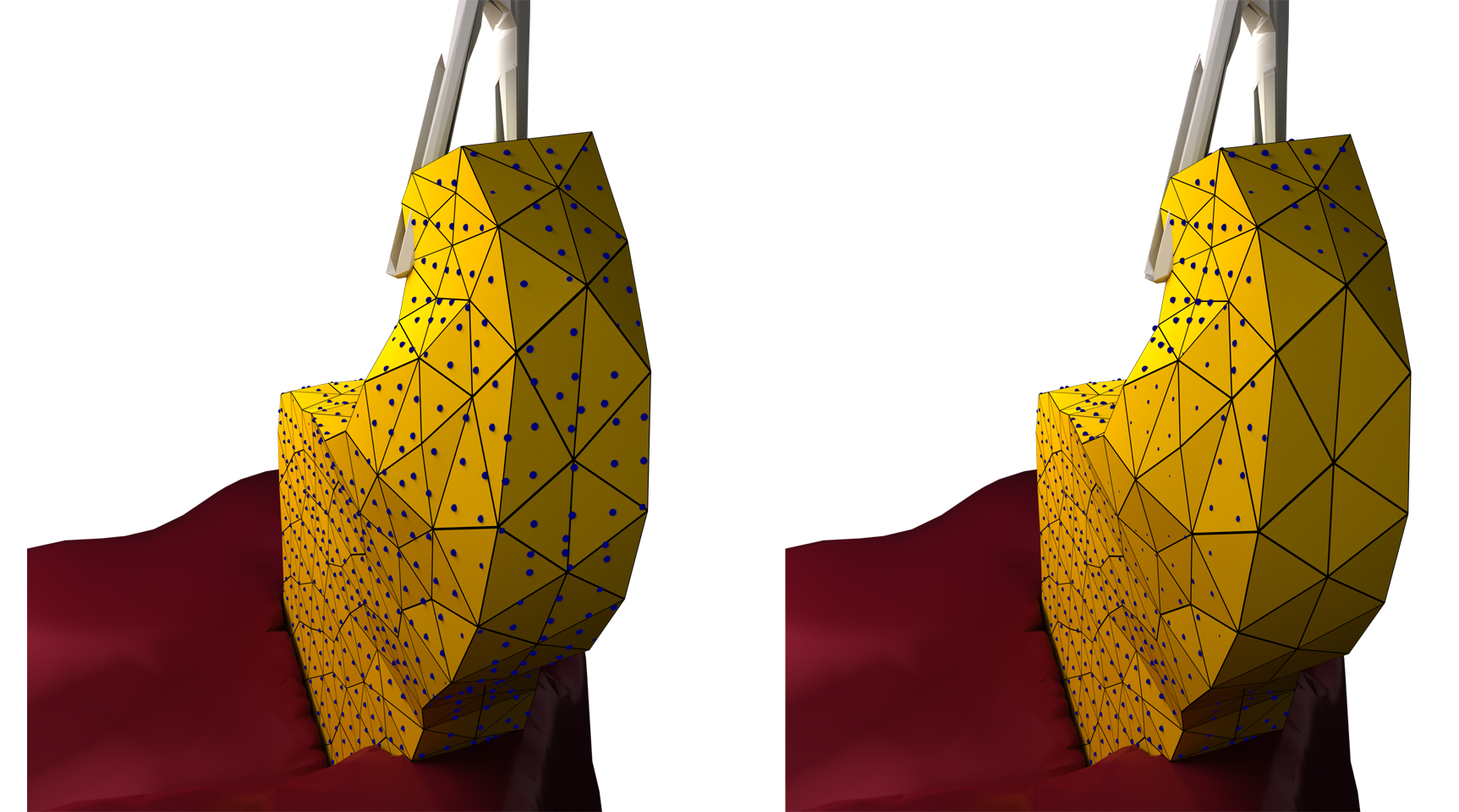}
	\caption{Overlay of the point cloud and the predicted mesh for two consecutive time steps $t=[70,71]$ in the Tissue Manipulation dataset. We repeat the point cloud from the earlier simulation step in both images for clarity. The illustration shows the correction behavior of \model by including the point cloud to ground the mesh based simulation in the time step.}
    \label{app_fig:figure1_tissue}
\end{figure}

\begin{figure}
    \centering
	\includegraphics[width=0.98\textwidth]{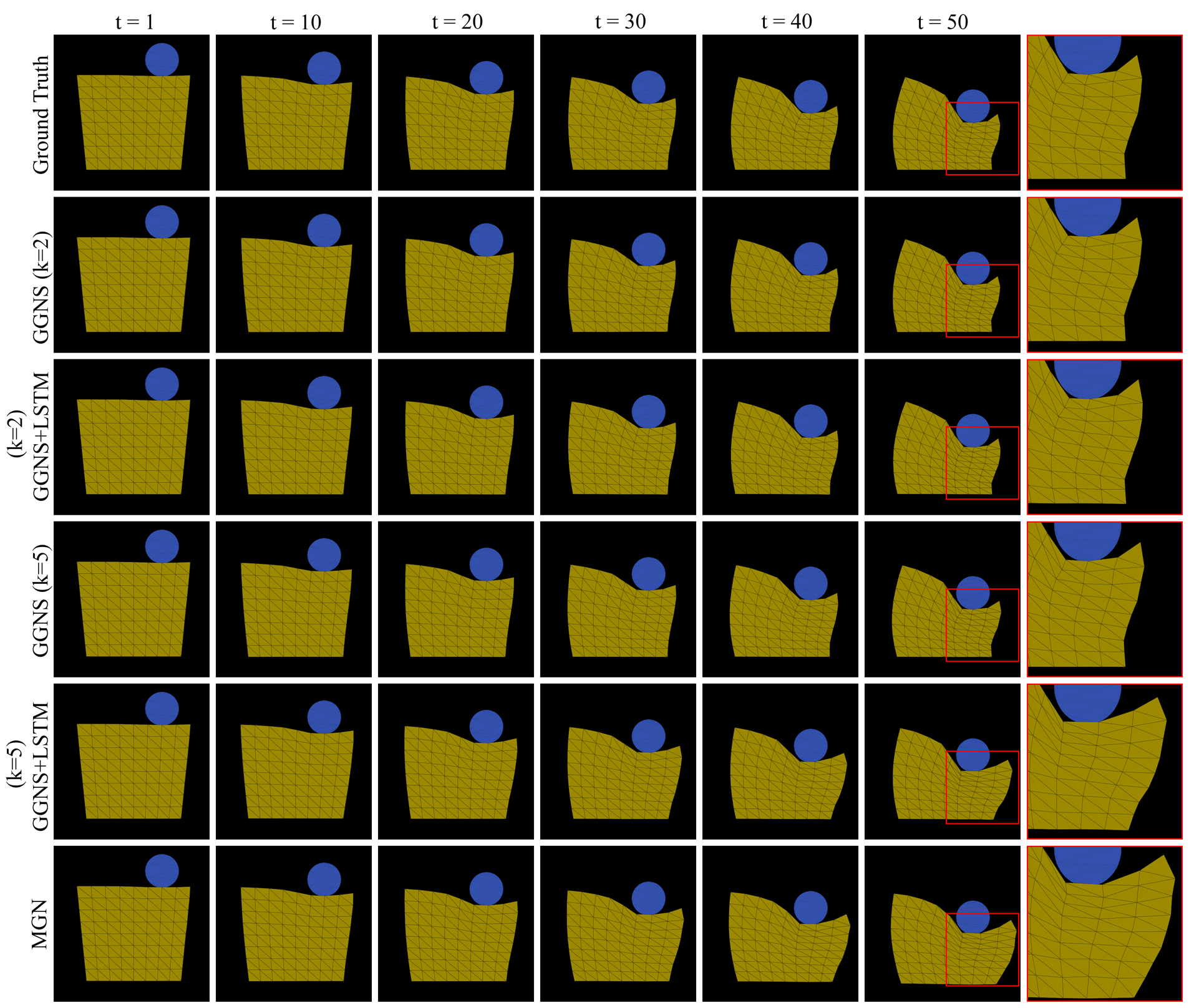}
	\caption{Test rollout visualization for the Deformable Plate task. The last column depicts a close-up of the final time step, which is shown in full in the previous column. Here, we additionally show qualitative results for the GGNS+LSTM model. We can see that for $k=2$ it matches the ground truth quite well, while for $k=5$ a large error occurs due to a prediction of the wrong material.}
    \label{app_fig:plate_stripe_qualitative}
    \vspace{0.25cm}
\end{figure}

\begin{figure}
    \centering
	\includegraphics[width=0.98\textwidth]{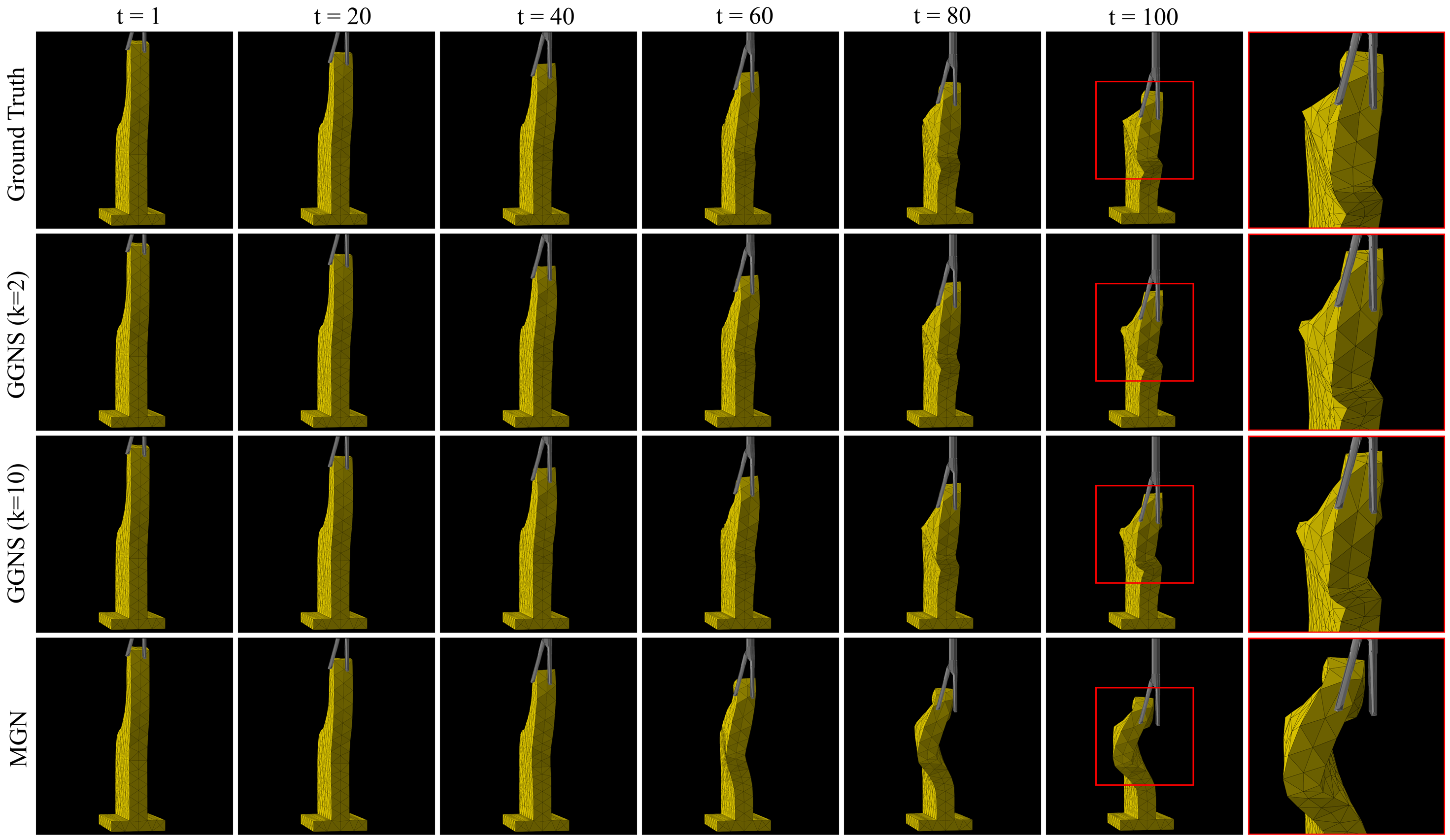}
	\caption{Test rollout visualization for the Tissue Manipulation task. The last column depicts a close-up of the final time step, which is shown in full in the previous column.}
    \label{app_fig:tissue_stripe_qualitative}
    \vspace{0.25cm}
\end{figure}

\begin{figure}
    \centering
	\includegraphics[width=0.98\textwidth]{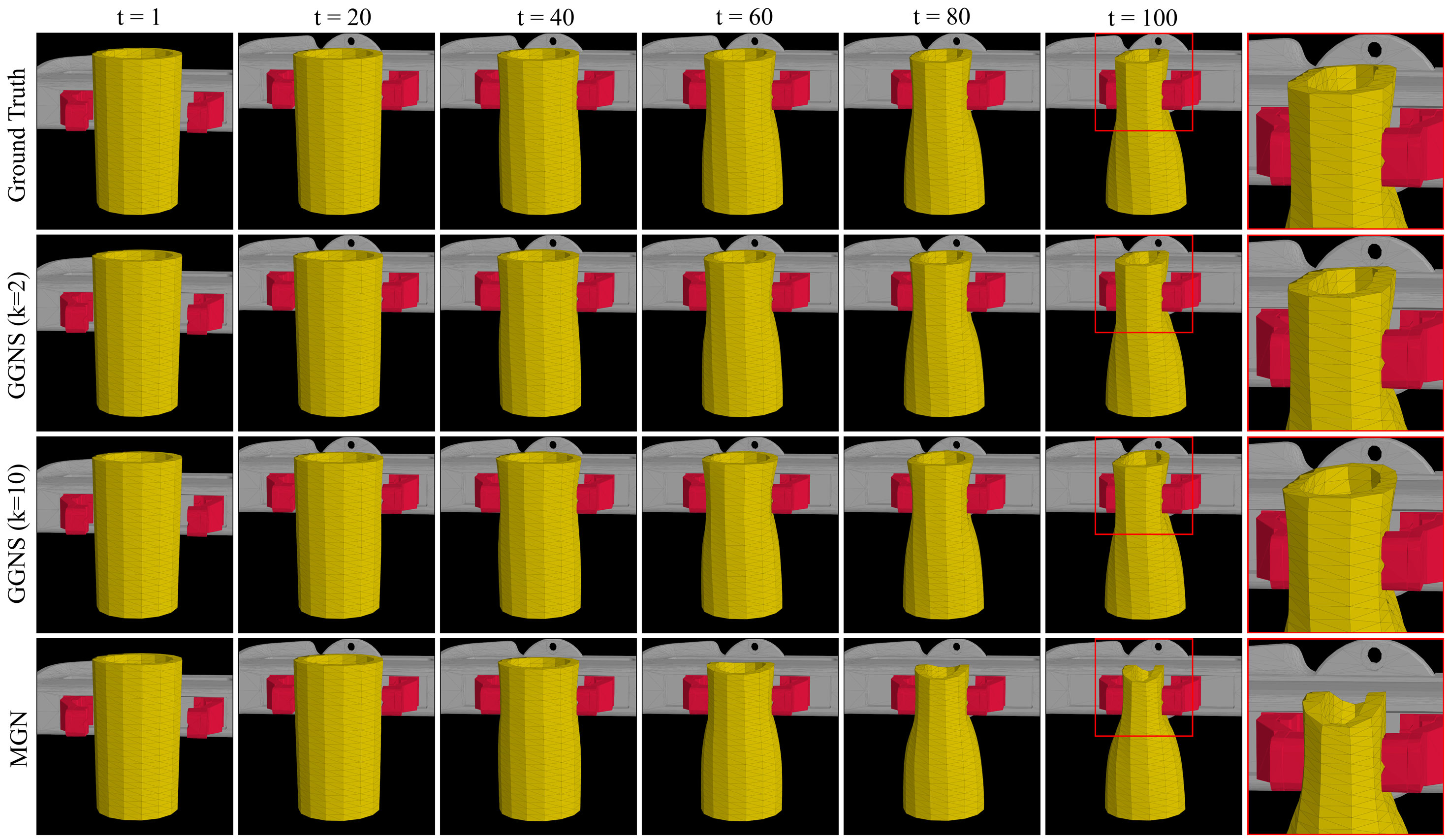}
	\caption{Test rollout visualization for the Cavity Grasping task. The last column depicts a close-up of the final time step, which is shown in full in the previous column.}
    \label{app_fig:tube_stripe_qualitative}
    \vspace{0.25cm}
\end{figure}

%\todo{we could consider adding the timesteps as an x-label for the visualizations} 
In addition to the qualitative illustrations in the main paper, we also provide further views and examples here: Figure~\ref{app_fig:tissue_qualitative_left} shows the same trajectory as Figure~\ref{fig:tissue_qualitative} but from three additional viewing angles.
Figure~\ref{app_fig:figure1_plate} and Figure~\ref{app_fig:figure1_tissue} show an overlay of the point cloud on the deformable object during the time step where the simulation is grounded by the point cloud.
This representation is comparable to  Figure~\ref{fig:figure_one} for the Cavity Grasping dataset.
Furthermore, we provide example visualizations for a test rollout over time for the Deformable Plate task in Figure~\ref{app_fig:plate_stripe_qualitative}, for the Tissue Manipulation task in Figure~\ref{app_fig:tissue_stripe_qualitative}, and for the Cavity Grasping in Figure~\ref{app_fig:tube_stripe_qualitative}.
Throughout all tasks, \model closely matches the ground truth simulation for the complete rollout, achieving close to optimal results when provided with frequent point cloud information ($k=2$).
Opposed to this, \gls{mgn} sometimes fails to predict the correct material, leading to poor predictions over time and large mismatches in the final system states.

\clearpage
\section{Ablations}
\label{app_sec:ablations} 
\subsection{Hyperparameter Choices}
\begin{figure}
    \centering
	\includegraphics[width=0.98\textwidth]{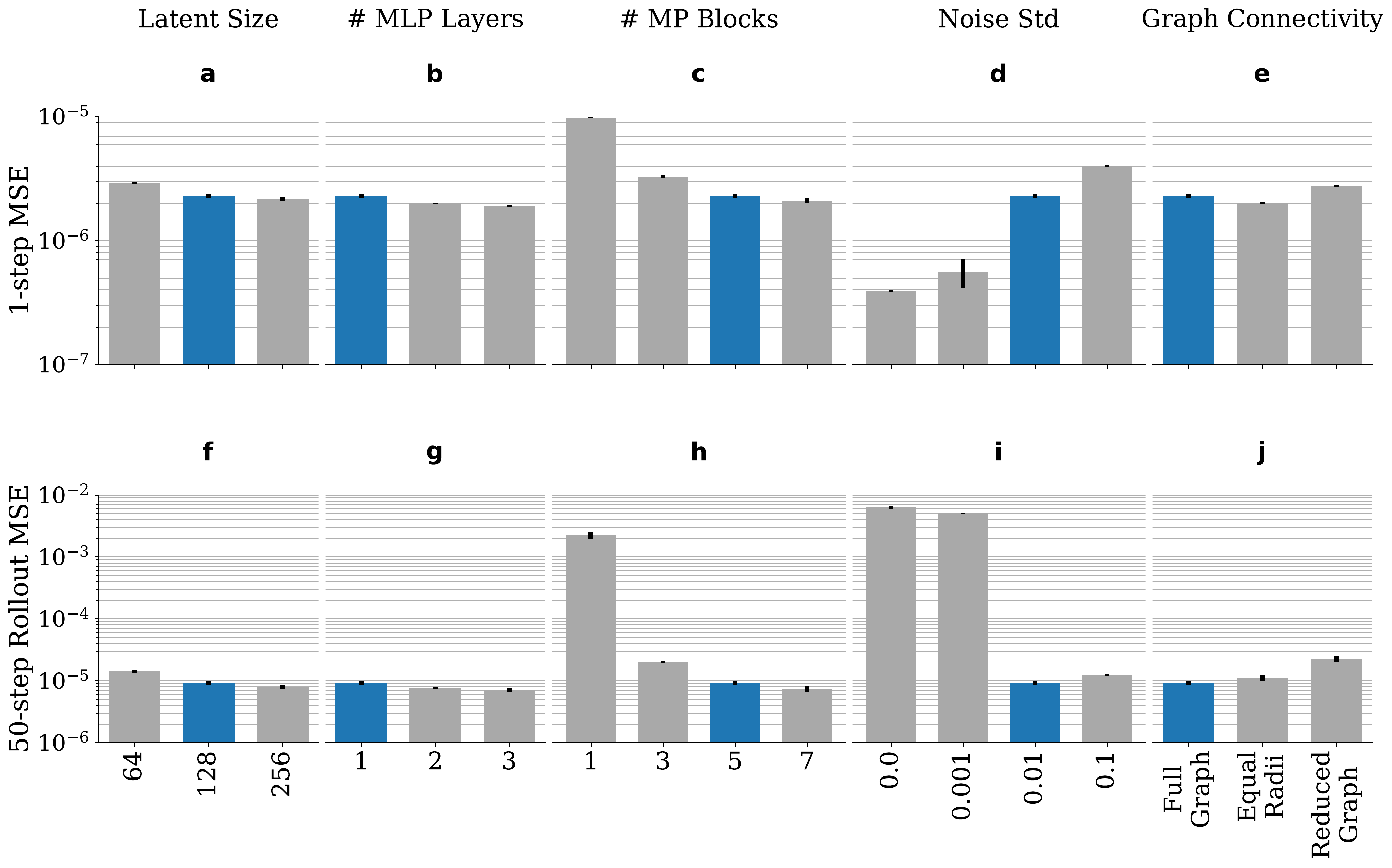}
	\caption{Performance for different changes in hyperparameter choices (grey) on the Deformable Plate dataset in comparison to our default model (blue) with $k=1$. Error bars indicate one standard deviation. The top row shows the error for the next-step prediction, the bottom row that of full rollouts.
	We find that a suitable noise scale is crucial for stable rollouts, and that more information in the form of additional edges between the different types of graph nodes generally improves performance. 
	Given enough Message Passing (MP) blocks, further increases in model capacity only lead to modest improvements.}
    \label{fig:ablations}
    \vspace{0.25cm}
\end{figure}

Figure~\ref{fig:ablations} compares the performance of \gls{ggns} for different hyperparameter choices. 
We find that the most importance parameters are the number of Message Passing (MP) blocks and the scale of the noise used in training. Both are crucial to achieve a good performance over multi-step rollouts. In terms of training noise, there is a $1$-step/multi-step loss trade-off. Other than that, our approach is robust to variations of the different hyperparameters. In terms of graph connectivity, it can be seen that all settings achieve similar performance. Additional information in the form of more local edges helps slightly, while larger connectivity radii do not do much. A detailed listing of the used edge radii is display in Table~\ref{tab:connectivity_2d}. In particular, the use of significantly more edges in the \textit{Equal Radii} setting does not provide a significant advantage, which is why we use weaker connectivity \textit{Full Graph} that saves computation time. The results for the \textit{Reduced Graph} settings show that edges within the point cloud are not mandatory. For this reason, we omit these edges in the more complex $3d$ tasks in favor of shorter computation time.

\subsection{Noisy Point Clouds}
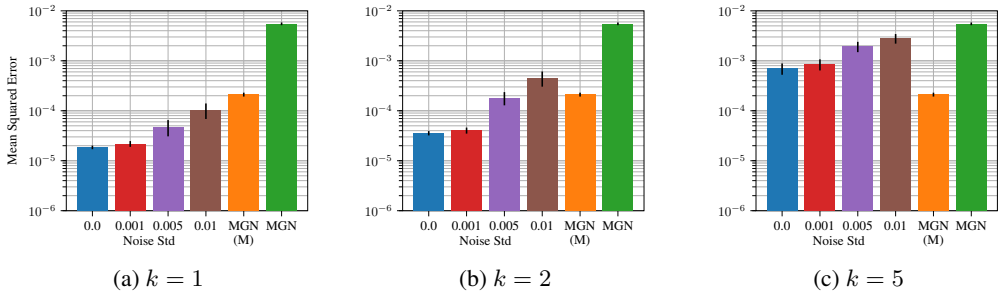
\begin{figure}
     \centering 
     \begin{subfigure}{.33\textwidth} 
         \centering 
         % This file was created with tikzplotlib v0.10.1.
\begin{tikzpicture}[font=\normalsize, scale=0.5]

\definecolor{crimson2143940}{RGB}{214,39,40}
\definecolor{darkgray176}{RGB}{176,176,176}
\definecolor{darkorange25512714}{RGB}{255,127,14}
\definecolor{forestgreen4416044}{RGB}{44,160,44}
\definecolor{mediumpurple148103189}{RGB}{148,103,189}
\definecolor{sienna1408675}{RGB}{140,86,75}
\definecolor{steelblue31119180}{RGB}{31,119,180}

\begin{axis}[
width=8cm,
log basis y={10},
tick align=outside,
tick pos=left,
x grid style={darkgray176},
xmajorgrids,
xmin=-0.69, xmax=5.69,
xtick style={color=black},
xtick={0,1,2,3,4,5},
xticklabels={0.0,
0.001,
0.005,
0.01,
MGN \\ (M),
MGN},
xticklabel style={align=center},
y grid style={darkgray176},
ylabel={Mean Squared Error},
ymajorgrids,
ymin=1e-06, ymax=0.01,
yminorgrids,
ymode=log,
extra x ticks ={1.5},
extra x tick labels={Noise Std},
extra x tick style={
    major tick length=1.4\baselineskip,
    major x tick style={draw=none},
    grid=none,
},
ytick style={color=black}
]

\draw[draw=none,fill=steelblue31119180] (axis cs:-0.4,1e-6) rectangle (axis cs:0.4,1.85631445160619e-05);

\draw[draw=none,fill=crimson2143940] (axis cs:0.6,1e-6) rectangle (axis cs:1.4,2.18389117055469e-05);

\draw[draw=none,fill=mediumpurple148103189] (axis cs:1.6,1e-6) rectangle (axis cs:2.4,4.79137478051362e-05);

\draw[draw=none,fill=sienna1408675] (axis cs:2.6,1e-6) rectangle (axis cs:3.4,0.000104276191746747);

\draw[draw=none,fill=darkorange25512714] (axis cs:3.6,1e-6) rectangle (axis cs:4.4,0.000211342016855876);

\draw[draw=none,fill=forestgreen4416044] (axis cs:4.6,1e-6) rectangle (axis cs:5.4,0.00550452694363064);

\path [draw=black, very thick]
(axis cs:0,1.70270005965326e-05)
--(axis cs:0,2.00992884355911e-05);

\path [draw=black, very thick]
(axis cs:1,1.8942337929366e-05)
--(axis cs:1,2.47354854817278e-05);

\path [draw=black, very thick]
(axis cs:2,3.07684747498691e-05)
--(axis cs:2,6.50590208604033e-05);

\path [draw=black, very thick]
(axis cs:3,6.85193085406013e-05)
--(axis cs:3,0.000140033074952893);

\path [draw=black, very thick]
(axis cs:4,0.000192350822438019)
--(axis cs:4,0.000230333211273732);

\path [draw=black, very thick]
(axis cs:5,0.00516381487138)
--(axis cs:5,0.00584523901588129);

\end{axis}

\end{tikzpicture}
         \caption{$k=1$}
         \label{fig:plate_noise_1}
     \end{subfigure}% <- noetig
     \begin{subfigure}{.33\textwidth} 
         \centering 
         % This file was created with tikzplotlib v0.10.1.
\begin{tikzpicture}[font=\normalsize, scale=0.5]

\definecolor{crimson2143940}{RGB}{214,39,40}
\definecolor{darkgray176}{RGB}{176,176,176}
\definecolor{darkorange25512714}{RGB}{255,127,14}
\definecolor{forestgreen4416044}{RGB}{44,160,44}
\definecolor{mediumpurple148103189}{RGB}{148,103,189}
\definecolor{sienna1408675}{RGB}{140,86,75}
\definecolor{steelblue31119180}{RGB}{31,119,180}

\begin{axis}[
width=8cm,
log basis y={10},
tick align=outside,
tick pos=left,
x grid style={darkgray176},
xmajorgrids,
xmin=-0.69, xmax=5.69,
xtick style={color=black},
xtick={0,1,2,3,4,5},
xticklabels={0.0,
0.001,
0.005,
0.01,
MGN \\ (M),
MGN},
xticklabel style={align=center},
y grid style={darkgray176},
ymajorgrids,
ymin=1e-06, ymax=0.01,
yminorgrids,
ymode=log,
extra x ticks ={1.5},
extra x tick labels={Noise Std},
extra x tick style={
    major tick length=1.4\baselineskip,
    major x tick style={draw=none},
    grid=none,
},
ytick style={color=black}
]

\draw[draw=none,fill=steelblue31119180] (axis cs:-0.4,1e-6) rectangle (axis cs:0.4,3.5661494952661e-05);

\draw[draw=none,fill=crimson2143940] (axis cs:0.6,1e-6) rectangle (axis cs:1.4,4.04813949708585e-05);

\draw[draw=none,fill=mediumpurple148103189] (axis cs:1.6,1e-6) rectangle (axis cs:2.4,0.000182375416049251);

\draw[draw=none,fill=sienna1408675] (axis cs:2.6,1e-6) rectangle (axis cs:3.4,0.000457162842927156);

\draw[draw=none,fill=darkorange25512714] (axis cs:3.6,1e-6) rectangle (axis cs:4.4,0.000211342016855876);

\draw[draw=none,fill=forestgreen4416044] (axis cs:4.6,1e-6) rectangle (axis cs:5.4,0.00550452694363064);

\path [draw=black, very thick]
(axis cs:0,3.21349572095228e-05)
--(axis cs:0,3.91880326957993e-05);

\path [draw=black, very thick]
(axis cs:1,3.4883191738637e-05)
--(axis cs:1,4.60795982030801e-05);

\path [draw=black, very thick]
(axis cs:2,0.000128465114282925)
--(axis cs:2,0.000236285717815576);

\path [draw=black, very thick]
(axis cs:3,0.000303907546287178)
--(axis cs:3,0.000610418139567133);

\path [draw=black, very thick]
(axis cs:4,0.000192350822438019)
--(axis cs:4,0.000230333211273732);

\path [draw=black, very thick]
(axis cs:5,0.00516381487138)
--(axis cs:5,0.00584523901588129);

\end{axis}

\end{tikzpicture}
         \caption{$k=2$}
         \label{fig:plate_noise_2}
     \end{subfigure} 
     \begin{subfigure}{.33\textwidth} 
         \centering 
         % This file was created with tikzplotlib v0.10.1.
\begin{tikzpicture}[font=\normalsize, scale=0.5]

\definecolor{crimson2143940}{RGB}{214,39,40}
\definecolor{darkgray176}{RGB}{176,176,176}
\definecolor{darkorange25512714}{RGB}{255,127,14}
\definecolor{forestgreen4416044}{RGB}{44,160,44}
\definecolor{mediumpurple148103189}{RGB}{148,103,189}
\definecolor{sienna1408675}{RGB}{140,86,75}
\definecolor{steelblue31119180}{RGB}{31,119,180}

\begin{axis}[
width=8cm,
log basis y={10},
tick align=outside,
tick pos=left,
x grid style={darkgray176},
xmajorgrids,
xmin=-0.69, xmax=5.69,
xtick style={color=black},
xtick={0,1,2,3,4,5},
xticklabels={0.0,
0.001,
0.005,
0.01,
MGN \\ (M),
MGN},
xticklabel style={align=center},
y grid style={darkgray176},
ymajorgrids,
ymin=1e-06, ymax=0.01,
yminorgrids,
ymode=log,
extra x ticks ={1.5},
extra x tick labels={Noise Std},
extra x tick style={
    major tick length=1.4\baselineskip,
    major x tick style={draw=none},
    grid=none,
},
ytick style={color=black}
]
\draw[draw=none,fill=steelblue31119180] (axis cs:-0.4,1e-6) rectangle (axis cs:0.4,0.000705036159797951);

\draw[draw=none,fill=crimson2143940] (axis cs:0.6,1e-6) rectangle (axis cs:1.4,0.00085463695526123);

\draw[draw=none,fill=mediumpurple148103189] (axis cs:1.6,1e-6) rectangle (axis cs:2.4,0.00194298259593822);

\draw[draw=none,fill=sienna1408675] (axis cs:2.6,1e-6) rectangle (axis cs:3.4,0.00281874498437952);
\draw[draw=none,fill=darkorange25512714] (axis cs:3.6,1e-6) rectangle (axis cs:4.4,0.000211342016855876);

\draw[draw=none,fill=forestgreen4416044] (axis cs:4.6,1e-6) rectangle (axis cs:5.4,0.00550452694363064);

\path [draw=black, very thick]
(axis cs:0,0.000525538879921704)
--(axis cs:0,0.000884533439674198);

\path [draw=black, very thick]
(axis cs:1,0.000640927497247471)
--(axis cs:1,0.00106834641327499);

\path [draw=black, very thick]
(axis cs:2,0.00148987862305591)
--(axis cs:2,0.00239608656882054);

\path [draw=black, very thick]
(axis cs:3,0.00220500223310386)
--(axis cs:3,0.00343248773565519);

\path [draw=black, very thick]
(axis cs:4,0.000192350822438019)
--(axis cs:4,0.000230333211273732);

\path [draw=black, very thick]
(axis cs:5,0.00516381487138)
--(axis cs:5,0.00584523901588129);

\end{axis}

\end{tikzpicture}
         \caption{$k=5$}
         \label{fig:plate_noise_5}
     \end{subfigure} 
     \caption{Additional ablations for more realistic point cloud data on two datasets. Here, four different noise levels on the point cloud are evaluated on the Deformable Plate datset. Different grounding frequencies of $k=1$ in (\subref{fig:plate_noise_1}), $k=2$ in (\subref{fig:plate_noise_2}) and $k=5$ in (\subref{fig:plate_noise_5}). \model performs better than the baseline even when noise in the scale of the training noise of $\sigma = 0.01$ is applied to the point cloud.}
     \label{app_fig:plate_noise}
\end{figure}

Besides the ablations on our hyperparameter choices, we present further ablations on more realistic point cloud data. 
For this purpose, we use point clouds with additional noise and only partial observability to get closer to real world point clouds. 
Figure~\ref{app_fig:plate_noise} shows the results for additional ablations on different scales of noise on the point cloud data of the Deformable Plate dataset.
We add noise to the point cloud positions during training, evaluation and testing.
This makes it more difficult to infer the correct behavior from the point cloud, but provides a more realistic scenario for, because real world point clouds often exhibit large noise. 
The results show the robustness of our method: Even when a noise level of $\sigma = 0.01$ is applied to the point cloud during testing, it clearly outperforms the baseline. 
This noise level corresponds to the amount of noise used on the mesh during training.

\subsection{Partial Observable Point Clouds}
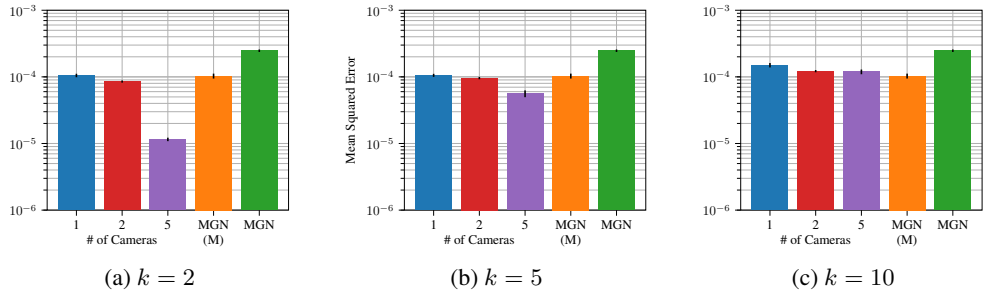
\begin{figure}
     \centering 
     % \begin{subfigure}{0.4\textwidth} 
     %     \centering 
     %     \input{tikz/appendix/1_hop_tube_bar_loss_complete_rollout.tex}
     %     \caption{$k=1$}
     %     \label{fig:tube_partially_1}
     % \end{subfigure}% <- noetig
     \begin{subfigure}{0.33\textwidth} 
         \centering 
         % This file was created with tikzplotlib v0.10.1.
\begin{tikzpicture}[font=\normalsize, scale=0.5]

\definecolor{crimson2143940}{RGB}{214,39,40}
\definecolor{darkgray176}{RGB}{176,176,176}
\definecolor{darkorange25512714}{RGB}{255,127,14}
\definecolor{forestgreen4416044}{RGB}{44,160,44}
\definecolor{mediumpurple148103189}{RGB}{148,103,189}
\definecolor{steelblue31119180}{RGB}{31,119,180}

\begin{axis}[
width=8.0cm,
log basis y={10},
tick align=outside,
tick pos=left,
x grid style={darkgray176},
xmajorgrids,
xmin=-0.64, xmax=4.64,
xtick style={color=black},
xtick={0,1,2,3,4},
xticklabels={1,
2,
5,
MGN \\ (M),
MGN},
xticklabel style={align=center},
y grid style={darkgray176},
ymajorgrids,
ymin=1e-06, ymax=0.001,
yminorgrids,
ymode=log,
extra x ticks ={1.0},
extra x tick labels={\# of Cameras},
extra x tick style={
    major tick length=1.4\baselineskip,
    major x tick style={draw=none},
    grid=none,
},
ytick style={color=black}
]
\draw[draw=none,fill=steelblue31119180] (axis cs:-0.4,1e-6) rectangle (axis cs:0.4,0.000105071063836416);

\draw[draw=none,fill=crimson2143940] (axis cs:0.6,1e-6) rectangle (axis cs:1.4,8.50927770137787e-05);

\draw[draw=none,fill=mediumpurple148103189] (axis cs:1.6,1e-6) rectangle (axis cs:2.4,1.15319627026717e-05);

\draw[draw=none,fill=darkorange25512714] (axis cs:2.6,1e-6) rectangle (axis cs:3.4,0.000103072641293208);

\draw[draw=none,fill=forestgreen4416044] (axis cs:3.6,1e-6) rectangle (axis cs:4.4,0.000246960852543513);

\path [draw=black, very thick]
(axis cs:0,9.8799545517818e-05)
--(axis cs:0,0.000111342582155013);

\path [draw=black, very thick]
(axis cs:1,8.19460851416647e-05)
--(axis cs:1,8.82394688858927e-05);

\path [draw=black, very thick]
(axis cs:2,1.08419677329353e-05)
--(axis cs:2,1.22219576724081e-05);

\path [draw=black, very thick]
(axis cs:3,9.43293592293738e-05)
--(axis cs:3,0.000111815923357042);

\path [draw=black, very thick]
(axis cs:4,0.000236388867833816)
--(axis cs:4,0.000257532837253209);

\end{axis}

\end{tikzpicture}
         \caption{$k=2$}
         \label{fig:tube_partially_2}
     \end{subfigure}%
     \begin{subfigure}{0.33\textwidth} 
         \centering 
         % This file was created with tikzplotlib v0.10.1.
\begin{tikzpicture}[font=\normalsize, scale=0.5]

\definecolor{crimson2143940}{RGB}{214,39,40}
\definecolor{darkgray176}{RGB}{176,176,176}
\definecolor{darkorange25512714}{RGB}{255,127,14}
\definecolor{forestgreen4416044}{RGB}{44,160,44}
\definecolor{mediumpurple148103189}{RGB}{148,103,189}
\definecolor{steelblue31119180}{RGB}{31,119,180}

\begin{axis}[
width=8.0cm,
log basis y={10},
tick align=outside,
tick pos=left,
x grid style={darkgray176},
xmajorgrids,
xmin=-0.64, xmax=4.64,
xtick style={color=black},
xtick={0,1,2,3,4},
xticklabels={1,
2,
5,
MGN \\ (M),
MGN},
xticklabel style={align=center},
y grid style={darkgray176},
ylabel={Mean Squared Error},
ymajorgrids,
ymin=1e-06, ymax=0.001,
yminorgrids,
ymode=log,
extra x ticks ={1.0},
extra x tick labels={\# of Cameras},
extra x tick style={
    major tick length=1.4\baselineskip,
    major x tick style={draw=none},
    grid=none,
},
ytick style={color=black}
]
\draw[draw=none,fill=steelblue31119180] (axis cs:-0.4,1e-6) rectangle (axis cs:0.4,0.000105518877506256);

\draw[draw=none,fill=crimson2143940] (axis cs:0.6,1e-6) rectangle (axis cs:1.4,9.5943487683932e-05);

\draw[draw=none,fill=mediumpurple148103189] (axis cs:1.6,1e-6) rectangle (axis cs:2.4,5.61035081744194e-05);

\draw[draw=none,fill=darkorange25512714] (axis cs:2.6,1e-6) rectangle (axis cs:3.4,0.000103072641293208);

\draw[draw=none,fill=forestgreen4416044] (axis cs:3.6,1e-6) rectangle (axis cs:4.4,0.000246960852543513);

\path [draw=black, very thick]
(axis cs:0,9.99390604907162e-05)
--(axis cs:0,0.000111098694521796);

\path [draw=black, very thick]
(axis cs:1,9.21195221701624e-05)
--(axis cs:1,9.97674531977015e-05);

\path [draw=black, very thick]
(axis cs:2,4.9541341389836e-05)
--(axis cs:2,6.26656749590027e-05);

\path [draw=black, very thick]
(axis cs:3,9.43293592293738e-05)
--(axis cs:3,0.000111815923357042);

\path [draw=black, very thick]
(axis cs:4,0.000236388867833816)
--(axis cs:4,0.000257532837253209);

\end{axis}

\end{tikzpicture}
         \caption{$k=5$}
         \label{fig:tube_partially_5}
     \end{subfigure}%
     \begin{subfigure}{0.33\textwidth} 
         \centering 
         % This file was created with tikzplotlib v0.10.1.
\begin{tikzpicture}[font=\normalsize, scale=0.5]

\definecolor{crimson2143940}{RGB}{214,39,40}
\definecolor{darkgray176}{RGB}{176,176,176}
\definecolor{darkorange25512714}{RGB}{255,127,14}
\definecolor{forestgreen4416044}{RGB}{44,160,44}
\definecolor{mediumpurple148103189}{RGB}{148,103,189}
\definecolor{steelblue31119180}{RGB}{31,119,180}

\begin{axis}[
width=8.0cm,
log basis y={10},
tick align=outside,
tick pos=left,
x grid style={darkgray176},
xmajorgrids,
xmin=-0.64, xmax=4.64,
xtick style={color=black},
xtick={0,1,2,3,4},
xticklabels={1,
2,
5,
MGN \\ (M),
MGN},
xticklabel style={align=center},
y grid style={darkgray176},
ymajorgrids,
ymin=1e-06, ymax=0.001,
yminorgrids,
ymode=log,
extra x ticks ={1.0},
extra x tick labels={\# of Cameras},
extra x tick style={
    major tick length=1.4\baselineskip,
    major x tick style={draw=none},
    grid=none,
},
ytick style={color=black}
]
ytick style={color=black}
]
\draw[draw=none,fill=steelblue31119180] (axis cs:-0.4,1e-6) rectangle (axis cs:0.4,0.000149875676631928);

\draw[draw=none,fill=crimson2143940] (axis cs:0.6,1e-6) rectangle (axis cs:1.4,0.000121978318691254);

\draw[draw=none,fill=mediumpurple148103189] (axis cs:1.6,1e-6) rectangle (axis cs:2.4,0.000119192347703774);

\draw[draw=none,fill=darkorange25512714] (axis cs:2.6,1e-6) rectangle (axis cs:3.4,0.000103072641293208);

\draw[draw=none,fill=forestgreen4416044] (axis cs:3.6,1e-6) rectangle (axis cs:4.4,0.000246960852543513);

\path [draw=black, very thick]
(axis cs:0,0.000138561042440151)
--(axis cs:0,0.000161190310823704);

\path [draw=black, very thick]
(axis cs:1,0.000117291474778529)
--(axis cs:1,0.000126665162603978);

\path [draw=black, very thick]
(axis cs:2,0.000109517221598376)
--(axis cs:2,0.000128867473809172);

\path [draw=black, very thick]
(axis cs:3,9.43293592293738e-05)
--(axis cs:3,0.000111815923357042);

\path [draw=black, very thick]
(axis cs:4,0.000236388867833816)
--(axis cs:4,0.000257532837253209);

\end{axis}

\end{tikzpicture}
         \caption{$k=10$}
         \label{fig:tube_partially_10}
     \end{subfigure}
     \caption{Additional ablations for more realistic point cloud data on the Cavity Grasping dataset. For this purpose, different numbers of cameras are used when generating the point cloud using raycasting. Comparison for three different grounding frequencies:$k=2$ in (\subref{fig:tube_partially_2}), $k=5$ in (\subref{fig:tube_partially_5}) and $k=10$ in (\subref{fig:tube_partially_10}). \model outperforms the baseline for all camera settings and grounding frequencies $k$.} 
     \label{app_fig:tube_partially}
\end{figure}

\begin{figure}
    \centering
	\includegraphics[width=1.0\textwidth]{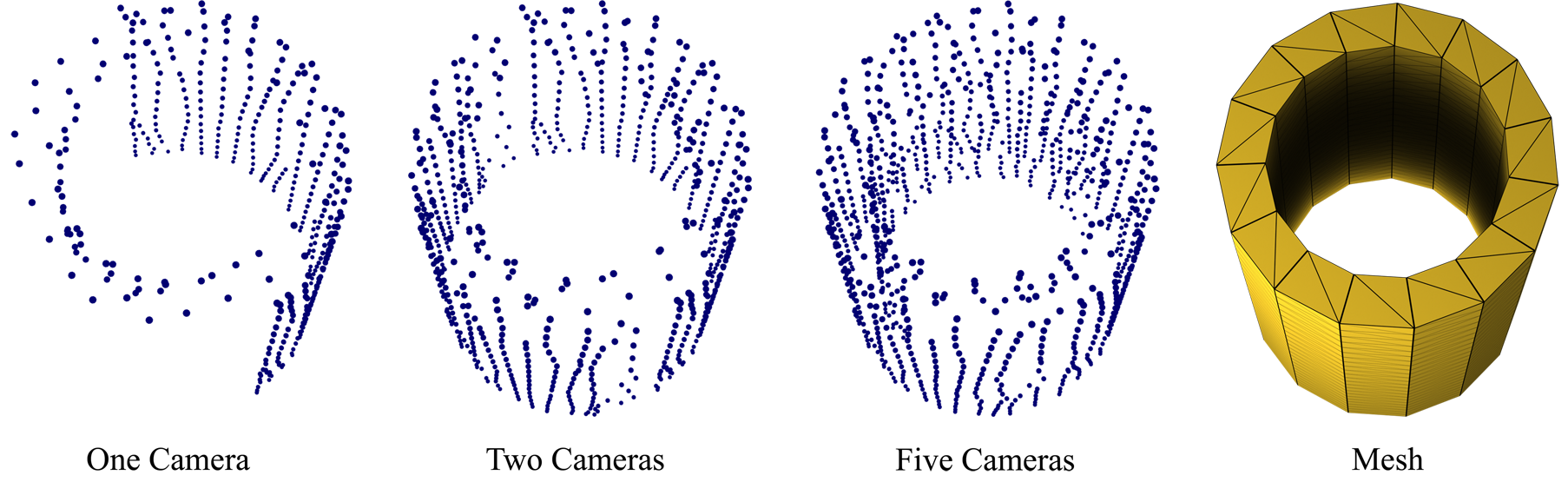}
	\caption{Visualization of the point clouds using one, two or five cameras for the raycasting and the corresponding mesh for reference. It is clearly visible how better coverage of the object is achieved as the number of cameras increases.}
    \label{app_fig:add_experiments_visual}
\end{figure}

For the ablations on the partial observability, we use the Cavity Grasping dataset.
We generate the partial point clouds by using only one, two or five virtual point cloud cameras when using raycasting. 
The resulting point clouds are visualized for better clarity in Figure~\ref{app_fig:add_experiments_visual} for an example test trajectory at time step $t=0$.
One camera results in a coverage from only one half of the outer surface of the cavity and two cameras cover almost the complete outer hull but not the inner surface. 
With five cameras, the point cloud covers almost the entire mesh completely, except for the inside and bottom. 
The resulting point clouds have a very different number of points: About $400$ for one camera, about $600$ for two cameras, and about $1000$ for five cameras compared to 750 mesh nodes for the cavity.
The results in Figure~\ref{app_fig:tube_partially} show that even with these much less complete point clouds, \model still outperforms the baseline. For $k \leq 5$ this is the case even if the baseline has access to the full initial state, which \model has not.

\newpage

% \begin{figure}
%      \centering 
%      \begin{subfigure}{.24\textwidth} 
%          \centering 
%          \includegraphics[width=1.0\textwidth]{iclr2023/figures/appendix/1camera_top_view.png}
%          \caption{One Camera}
%          \label{app_fig:add_experiments_visual_1}
%      \end{subfigure}% <- noetig
%      \begin{subfigure}{.24\textwidth} 
%          \centering 
%          \includegraphics[width=1.0\textwidth]{iclr2023/figures/appendix/2camera_top_view.png}
%          \caption{Two Cameras}
%          \label{app_fig:add_experiments_visual_2}
%      \end{subfigure} 
%      \begin{subfigure}{.24\textwidth} 
%          \centering 
%          \includegraphics[width=1.0\textwidth]{iclr2023/figures/appendix/5camera_top_view.png}
%          \caption{Five Cameras}
%          \label{app_fig:add_experiments_visual_5}
%      \end{subfigure} 
%      \begin{subfigure}{.24\textwidth} 
%          \centering 
%          \includegraphics[width=1.0\textwidth]{iclr2023/figures/appendix/mesh_top_view.png}
%          \caption{Mesh}
%          \label{app_fig:add_experiments_visual_5}
%      \end{subfigure}
%      \caption{Visualization of the point clouds using one (\subref{app_fig:add_experiments_visual_1}), two (\subref{app_fig:add_experiments_visual_2}) or five (\subref{app_fig:add_experiments_visual_5}) cameras for the raycasting. It is clearly visible how better coverage of the object is achieved as the number of cameras increases.}
%      \label{app_fig:add_experiments_visual}
% \end{figure} 

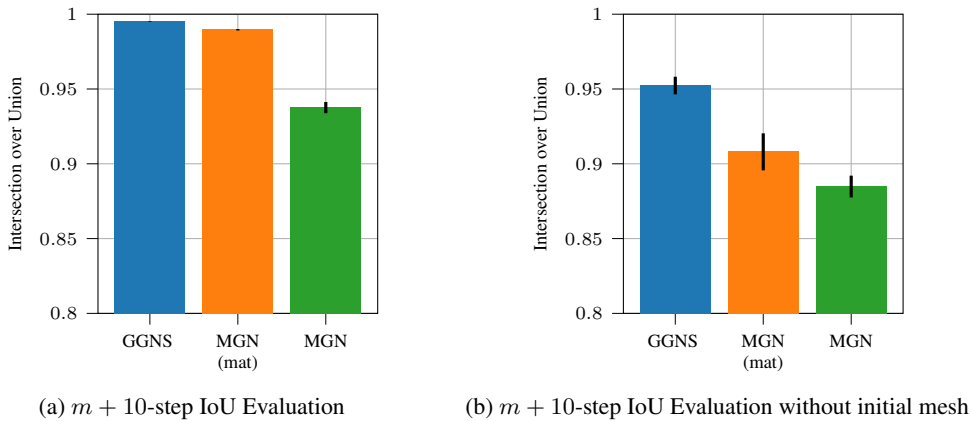
\begin{figure}
     \centering 
     \begin{subfigure}{.5\textwidth} 
         \centering 
         % This file was created with tikzplotlib v0.10.1.
\begin{tikzpicture}[font=\scriptsize]

\definecolor{crimson2143940}{RGB}{214,39,40}
\definecolor{darkgray176}{RGB}{176,176,176}
\definecolor{darkorange25512714}{RGB}{255,127,14}
\definecolor{forestgreen4416044}{RGB}{44,160,44}
\definecolor{steelblue31119180}{RGB}{31,119,180}

\begin{axis}[
width=5.3cm,
height=5.562cm,
tick align=outside,
tick pos=left,
x grid style={darkgray176},
xmajorgrids,
xmin=-0.59, xmax=2.59,
xtick style={color=black},
xtick={0,1,2,3},
xticklabels={\model, 
MGN\\(mat), 
MGN},
xticklabel style={align=center},
y grid style={darkgray176},
ylabel={Intersection over Union},
ymajorgrids,
ymin=0.8, ymax=1,
ytick style={color=black}
]
\draw[draw=none,fill=steelblue31119180] (axis cs:-0.4,0) rectangle (axis cs:0.4,0.995120806663238);

\draw[draw=none,fill=darkorange25512714] (axis cs:0.6,0) rectangle (axis cs:1.4,0.989616376192604);

\draw[draw=none,fill=forestgreen4416044] (axis cs:1.6,0) rectangle (axis cs:2.4,0.937550472929688);

\path [draw=black, very thick]
(axis cs:0,0.994892927386393)
--(axis cs:0,0.995348685940083);

\path [draw=black, very thick]
(axis cs:1,0.989114434582318)
--(axis cs:1,0.990118317802889);

\path [draw=black, very thick]
(axis cs:2,0.93380151322258)
--(axis cs:2,0.941299432636796);

\end{axis}

\end{tikzpicture}
         \caption{$m+10$-step IoU Evaluation}
         \label{fig:trapez_bar_IOU_mpc_10}
     \end{subfigure}% <- noetig
     \begin{subfigure}{.5\textwidth} 
         \centering 
         % This file was created with tikzplotlib v0.10.1.
\begin{tikzpicture}[font=\scriptsize] %\footnotesize

\definecolor{crimson2143940}{RGB}{214,39,40}
\definecolor{darkgray176}{RGB}{176,176,176}
\definecolor{darkorange25512714}{RGB}{255,127,14}
\definecolor{forestgreen4416044}{RGB}{44,160,44}
\definecolor{steelblue31119180}{RGB}{31,119,180}

\begin{axis}[
width=5.3cm,
height=5.562cm,
tick align=outside,
tick pos=left,
x grid style={darkgray176},
xmajorgrids,
xmin=-0.59, xmax=2.59,
xtick style={color=black},
xtick={0,1,2,3},
xticklabels={\model, 
MGN\\(mat), 
MGN},
xticklabel style={align=center},
y grid style={darkgray176},
ylabel={Intersection over Union},
ymajorgrids,
ymin=0.8, ymax=1,
ytick style={color=black}
]
\draw[draw=none,fill=steelblue31119180] (axis cs:-0.4,0) rectangle (axis cs:0.4,0.952306410886259);

\draw[draw=none,fill=darkorange25512714] (axis cs:0.6,0) rectangle (axis cs:1.4,0.907986720825382);

\draw[draw=none,fill=forestgreen4416044] (axis cs:1.6,0) rectangle (axis cs:2.4,0.884780660412549);

\path [draw=black, very thick]
(axis cs:0,0.946395237259766)
--(axis cs:0,0.958217584512751);

\path [draw=black, very thick]
(axis cs:1,0.895608205781345)
--(axis cs:1,0.920365235869419);

\path [draw=black, very thick]
(axis cs:2,0.877485826354636)
--(axis cs:2,0.892075494470463);

\end{axis}

\end{tikzpicture}
         \caption{$m+10$-step IoU Evaluation without initial mesh} 
         \label{fig:automesh_bar_IoU_mpc_10}
     \end{subfigure} 
     \caption{(\subref{fig:trapez_bar_IOU_mpc_10}) Comparison of our model to the baseline results on the Plate cataset using the $m+10$-step Evaluation routine. (\subref{fig:automesh_bar_IoU_mpc_10}) Results when using an initial mesh generated from the point cloud. \model outperforms the \gls{mgn} baseline even if it has access to the initial ground truth mesh.} 
\end{figure}

\begin{figure}
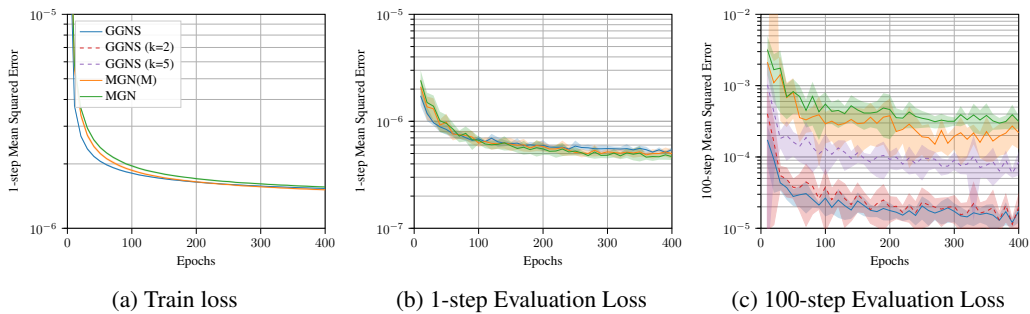

     \centering 
     \begin{subfigure}{.33\textwidth} 
         \centering         
         \input{tikz/appendix/tube_train_loss.tex}
            \caption{Train loss} 
            \label{fig:learning_curves_train_loss}
     \end{subfigure}% 
     \begin{subfigure}{.33\textwidth} 
         \centering 
         \input{tikz/appendix/tube_eval_loss_single_step.tex}
            \caption{1-step Evaluation Loss}
            \label{fig:learning_curves_single_step}
     \end{subfigure}% <- noetig
     \begin{subfigure}{.33\textwidth} 
         \centering 
         \input{tikz/appendix/tube_eval_loss_complete_rollout.tex}
            \caption{100-step Evaluation Loss}
            \label{fig:learning_curves_complete_rollout}
     \end{subfigure}% <- noetig    
     \caption{Exemplary learning curves for the Cavity Grasping task. The light shaded area indicates one standard deviation. Both \model and the baselines learn the task pretty similarly in terms of $1$-step predictions. Our model is only evaluated for the $k=2$ and $k=5$ variant during full rollout evaluation. Here, we can clearly see the advantage of using the point cloud information.} 
     \label{fig:learning_curves}
\end{figure}

\begin{table}
\caption{Edge radii for the connectivities between point clouds $\mathcal{P}$ and meshes $\mathcal{M}$ on the 2D Deformable Plate Dataset.}
\vspace*{0.5cm}
\centering
\begin{tabular}{llll}
\toprule
Setting & $\mathcal{P}-\mathcal{P}$ & $\mathcal{M}-\mathcal{P}$ & World\\
\midrule 
Full Graph & 0.1  & 0.08 & -\\
Equal Radii & 0.2 & 0.2 & - \\
Reduced Graph & 0.0 & 0.08  & -\\
MGN & 0.0 & 0.0 & 0.35\\
\hline
\end{tabular}
\label{tab:connectivity_2d}
\end{table}

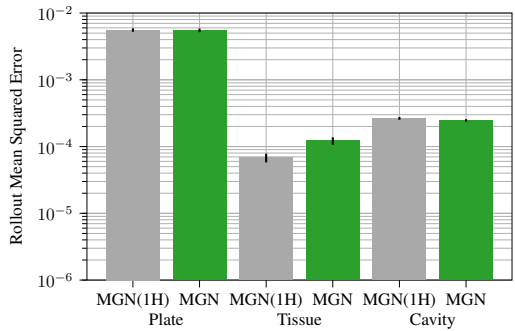
\begin{figure}
    \centering 
    \resizebox{0.5\textwidth}{!}{%        
        % This file was created with tikzplotlib v0.10.1.
\begin{tikzpicture}[font=\normalsize, scale=0.5]

\definecolor{crimson2143940}{RGB}{214,39,40}
\definecolor{darkgray176}{RGB}{176,176,176}
\definecolor{darkorange25512714}{RGB}{255,127,14}
\definecolor{forestgreen4416044}{RGB}{44,160,44}
\definecolor{mediumpurple148103189}{RGB}{148,103,189}
\definecolor{sienna1408675}{RGB}{140,86,75}
\definecolor{steelblue31119180}{RGB}{31,119,180}
\definecolor{gray169}{RGB}{169,169,169}

\begin{axis}[
width=10.2cm,
height=7cm,
log basis y={10},
tick align=outside,
tick pos=left,
x grid style={darkgray176},
xmajorgrids,
xmin=-0.69, xmax=5.69,
xtick style={color=black},
xtick={0,1,2,3,4,5},
xticklabels={
MGN(1H),
MGN,
MGN(1H),
MGN,
MGN(1H),
MGN},
xticklabel style={align=center},
y grid style={darkgray176},
ylabel={Rollout Mean Squared Error},
ymajorgrids,
ymin=1e-06, ymax=0.01,
yminorgrids,
ymode=log,
extra x ticks ={0.5, 2.5, 4.5},
extra x tick labels={Plate, Tissue, Cavity},
extra x tick style={
    major tick length=1.4\baselineskip,
    major x tick style={draw=none},
    grid=none,
},
ytick style={color=black}
]
\draw[draw=none,fill=gray169] (axis cs:-0.4,1e-06) rectangle (axis cs:0.4,0.0055708650942202);

\draw[draw=none,fill=forestgreen4416044] (axis cs:0.6,1e-06) rectangle (axis cs:1.4,0.00550452694363064);

\draw[draw=none,fill=gray169] (axis cs:1.6,1e-06) rectangle (axis cs:2.4,6.7988724509875e-05);
\draw[draw=none,fill=forestgreen4416044] (axis cs:2.6,1e-06) rectangle (axis cs:3.4,0.000121950203180313);
\draw[draw=none,fill=gray169] (axis cs:3.6,1e-06) rectangle (axis cs:4.4,0.000264605454603831);
\draw[draw=none,fill=forestgreen4416044] (axis cs:4.6,1e-06) rectangle (axis cs:5.4,0.000246960852543513);
\path [draw=black, very thick]
(axis cs:0,0.00524432136464868)
--(axis cs:0,0.00589740882379171);

\path [draw=black, very thick]
(axis cs:1,0.00516381487138)
--(axis cs:1,0.00584523901588129);

\path [draw=black, very thick]
(axis cs:2,5.78538892222831e-05)
--(axis cs:2,7.81235597974668e-05);

\path [draw=black, very thick]
(axis cs:3,0.000106454943636323)
--(axis cs:3,0.000137445462724304);

\path [draw=black, very thick]
(axis cs:4,0.000252039437275171)
--(axis cs:4,0.000277171471932491);

\path [draw=black, very thick]
(axis cs:5,0.000236388867833816)
--(axis cs:5,0.000257532837253209);

\end{axis}

\end{tikzpicture}
    }%    
    \caption{Comparison of the \gls{mgn} baseline with a version using the one-hot encoded edge types instead of an explicit edge type partitioning indicated by \textit{\gls{mgn} (1H)}. Both are compared for all three tasks and no significant advantage of the explicit edges partitioning could be found. For this reason, \model uses the one-hot encoding, because it is both conceptually simpler and requires less computational power. The \gls{mgn} baseline still uses explicit edge type partitioning throughout this work, following \cite{pfaff2020learning}.} 
    \label{fig:mgn_world_comparison_bar_loss_complete_rollout}
\end{figure}

\clearpage
\section{Hyperparameters}
\label{app_sec:hyperparameters} 
Table~\ref{tab:hyperparameters} gives an overview of hyperparameters shared across tasks.
Since \gls{gns} are generally robust to the choice of hyperparameters (c.f. \ref{app_sec:ablations}), we use the same hyperparameters for all task and for both, \model and \gls{mgn} for simplicity. 
The only hyperparameters that vary over tasks are the graph connectivity and the number of training epochs, as shown in Table~\ref{tab:task_hyperparameters}. 
We adapt these parameters to control for the total training time on a single GPU.

\begin{table}
\caption{Configuration of the hyperparameters and key information of the training of our model for all experiments.}
\vspace*{0.5cm}
\centering
\begin{tabular}{ll}
\toprule
Parameter & Value  \\
\midrule 
Batch Size & $32$ \\
Optimizer & Adam \\
Learning Rate & $\num{5e-4}$\\
Activation Function & LeakyReLU\\
Aggregation Function & Mean \\
Encoder & Linear Layer \\
MP-Blocks & $5$ \\
\gls{mlp} Layers & $1$ \\
Latent Dimension & $128$ \\
Decoder & $1$-layer \gls{mlp}\\
Residuals Connections & Around each MP block\\
Training Noise Std & 0.01\\
\bottomrule
\end{tabular}
\label{tab:hyperparameters}
\end{table}

\begin{table}
\caption{Task specific configuration and hyperparameters for our experiments. We vary the graph connectivity and the number of training epochs for different tasks to control the total training time of our method.}
\vspace*{0.5cm}
\centering
\begin{tabular}{llll}
\toprule
Parameter & Plate & Tissue & Cavity \\
\midrule 
Connectivity Setting & Full Graph & Reduced & Reduced\\
Number of Epochs & 1000 & 800 & 400\\
Approx. Training Time & $21:00~\si{\hour}$ & $40:00~\si{\hour}$ & $38:00~\si{\hour}$\\
\bottomrule
\end{tabular}
\label{tab:task_hyperparameters}
\end{table}

\end{document}